\documentclass[journal]{IEEEtran}
\usepackage{amsmath,amsfonts}
\usepackage{algorithmic}
\usepackage{algorithm}
\usepackage{array}
\usepackage{textcomp}
\usepackage{stfloats}
\usepackage{url}
\usepackage{verbatim}
\usepackage{graphicx}
\usepackage{cite}
\usepackage{subfigure}
\usepackage{comment}
\usepackage{wrapfig}
\usepackage{nicematrix}
\usepackage{multicol}
\usepackage{multirow}
\usepackage{booktabs}
\usepackage{colortbl}

\usepackage{xspace}
\newcommand{\abb}{\texttt{ProtoN-FM}\xspace}
\newcommand{\layer}{\textsl{ProtoNorm}\xspace}

\begin{document}

\title{Bridging Distribution Gaps in Time Series Foundation Model Pretraining with Prototype-Guided Normalization}

\author{Peiliang Gong, Emadeldeen Eldele, Min Wu, \IEEEmembership{Senior Member,~IEEE,} Zhenghua Chen, \IEEEmembership{Senior Member,~IEEE,} Xiaoli Li, \IEEEmembership{Fellow,~IEEE,} Daoqiang Zhang, \IEEEmembership{Senior Member,~IEEE}

\IEEEcompsocitemizethanks{
    \IEEEcompsocthanksitem Peiliang Gong and Daoqiang Zhang are with the Key Laboratory of Brain-Machine Intelligence Technology, Ministry of Education, College of Artificial Intelligence, Nanjing University of Aeronautics and Astronautics, Nanjing 211106, China 
    \IEEEcompsocthanksitem Emadeldeen Eldele, Zhenghua Chen, and Xiaoli Li are with the Institute for Infocomm Research (I$^2$R) and the Centre for Frontier AI Research (CFAR), Agency for Science, Technology and Research (A$*$STAR), Singapore 
    \IEEEcompsocthanksitem Min Wu is with the Institute for Infocomm Research (I$^2$R), Agency for Science, Technology and Research (A$*$STAR), Singapore 
}
}



\maketitle

\begin{abstract}
Foundation models have achieved remarkable success across diverse machine-learning domains through large-scale pretraining on large, diverse datasets. However, pretraining on such datasets introduces significant challenges due to substantial mismatches in data distributions, a problem particularly pronounced with time series data. In this paper, we tackle this issue by proposing a domain-aware adaptive normalization strategy within the Transformer architecture. Specifically, we replace the traditional LayerNorm with a prototype-guided dynamic normalization mechanism (\layer), where learned prototypes encapsulate distinct data distributions, and sample-to-prototype affinity determines the appropriate normalization layer. This mechanism effectively captures the heterogeneity of time series characteristics, aligning pretrained representations with downstream tasks. Through comprehensive empirical evaluation, we demonstrate that our method significantly outperforms conventional pretraining techniques across both classification and forecasting tasks, while effectively mitigating the adverse effects of distribution shifts during pretraining. Incorporating \layer is as simple as replacing a single line of code. Extensive experiments on diverse real-world time series benchmarks validate the robustness and generalizability of our approach, advancing the development of more versatile time series foundation models.
\end{abstract}


\section{Introduction}
\IEEEPARstart{F}{oundation} Models (FM) have revolutionized machine learning by learning general-purpose representations from vast amounts of unlabeled data \cite{FM_survey}. These models have achieved remarkable success, particularly in natural language processing (NLP) tasks \cite{BERT}. Prominent examples, such as GPT-3 \cite{GPT3}, GPT-4 \cite{GPT4}, and LLAMA \cite{touvron2023llama}, demonstrate exceptional performance and generalization capabilities, leveraging the intrinsic structures and patterns in textual data.

The potential of FMs to generalize across diverse domains holds immense promise for extending their efficacy to time series (TS) domains, such as finance \cite{yu2023temporal}, healthcare \cite{moor2023foundation}, and climate forecasting \cite{ts_weather}. However, unlike NLP tasks, where data distributions exhibit relative consistency and models can effectively capture patterns and semantics, training FMs on TS data presents a significant challenge due to mismatched pretraining data distributions \cite{RevIN}.
 
\begin{figure}[t]
    \centering
    \subfigure[]{\includegraphics[width=0.48\linewidth]{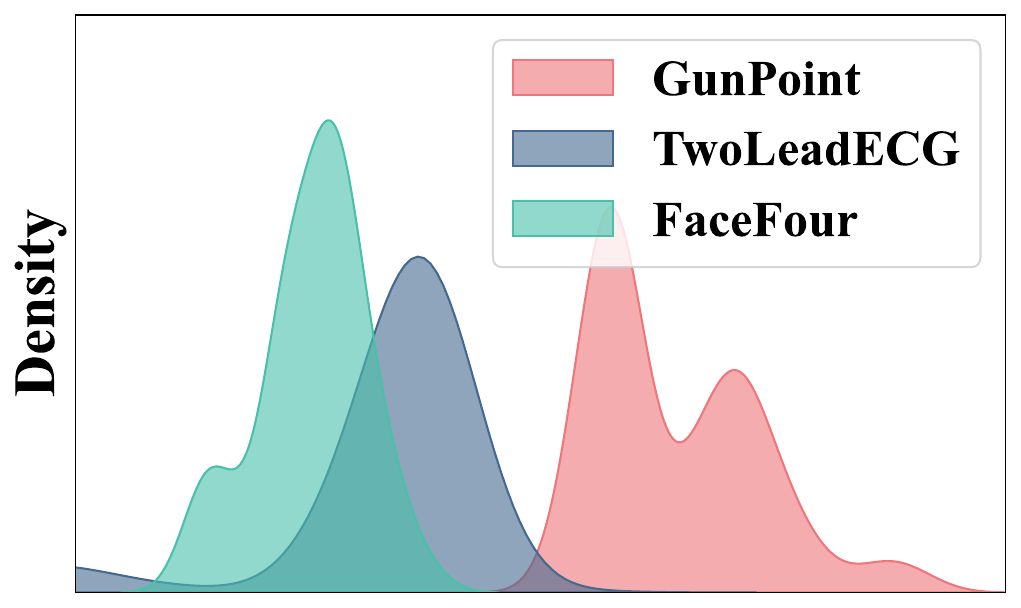}
    \label{figa:motivation_distribution}}
    \subfigure[]{\includegraphics[width=0.48\linewidth]{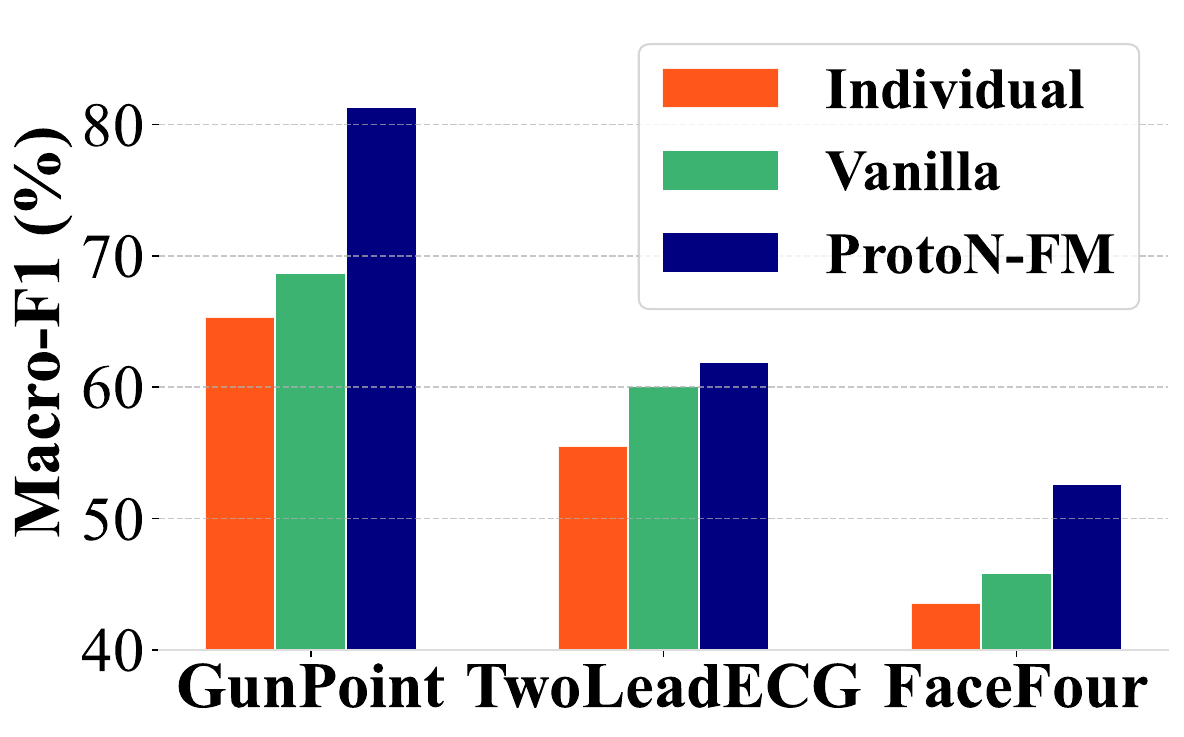}
    \label{figb:motivation_performance}}
    \caption{(a) Distributional shifts exist among three UCR time series datasets. (b) Fine-tuning performance comparison on these datasets after different pretraining strategies. \textit{Individual} refers to pretraining and fine-tuning a Transformer model on each dataset separately. \textit{Vanilla} denotes pretraining the foundation model on multiple datasets without additional design considerations. In \abb, we utilize the same multi-dataset pretraining, but incorporate our prototype-guided dynamic normalization mechanism, resulting in superior performance across diverse datasets.}
    \label{fig:motivation}
\end{figure}

This mismatch can be attributed to several factors. First, different TS data often manifest distinct properties, such as temporal dependencies, irregularities, and domain-specific dynamics. Second, TS data may exhibit varying sampling rates, number of channels, and noise levels, which deviate substantially from the clean and well-structured data employed in pretraining language models \cite{TS_survey, FM_ts_survey}. To illustrate this disparity, Figure \ref{figa:motivation_distribution} depicts the distribution of different types of datasets (i.e., GunPoint, TwoLeadECG, and FaceFour) in the UCR classification archive. These datasets, originating from diverse domains, exhibit significant variations in their value ranges and morphological characteristics, underscoring the inherent heterogeneity present in TS data.

The implications of this distributional mismatch on pretraining FMs are demonstrated in Figure~\ref{figb:motivation_performance}, which compares the fine-tuning performance across different pretraining strategies. The \textit{Vanilla} pretraining strategy, which neglects dataset heterogeneity, yields suboptimal fine-tuning results. Conversely, strategies that address this mismatch during pretraining demonstrate superior performance, highlighting the criticality of aligning FMs with the intrinsic characteristics of TS data.

In this work, we propose a novel approach to address the discrepancy between FM pretraining and TS data distributions. Specifically, we introduce an FM design based on a \textbf{Proto}type-guided dynamic \textbf{Norm}alization (\layer) mechanism within the Transformer architecture, enabling adaptive normalization tailored to the heterogeneous characteristics of TS data. As shown in Figure~\ref{fig:framework}, the \layer mechanism operates by learning a set of prototypes during pretraining, where each prototype encapsulates a cluster of data with shared attributes, such as temporal dependencies, noise levels, or sampling rates.

Unlike conventional LayerNorm, which applies fixed normalization parameters across all samples, our method assigns each prototype to its corresponding LayerNorm module, capturing the distinctive statistical properties of different data clusters. During training, the model computes an affinity score between each sample and the learned prototypes, dynamically selecting the most appropriate LayerNorm module based on the highest affinity. This adaptive selection process ensures that the normalization process aligns with the specific characteristics of each sample, effectively addressing the inherent heterogeneity in TS data.

Moreover, this adaptive mechanism integrates seamlessly within the Transformer architecture, leveraging its self-attention mechanism to further contextualize the prototype selection. By aligning the model's normalization process with the diverse characteristics of TS data, \layer mitigates the distribution shift during time series foundation model pretraining. This alignment enables the FM to generalize more effectively across domains and tasks, achieving superior fine-tuning performance and enhanced robustness compared to conventional approaches. The contributions of this work can be summarized as follows:
\begin{itemize}
    \item This is the first work to identify the critical challenge of data distribution mismatch between foundation model pretraining and time series data, which impedes the effective deployment of FMs to time series tasks.
    
    \item We propose a prototype-guided dynamic normalization mechanism (\layer) that adaptively normalizes features based on learned distributional prototypes. This lightweight and modular solution integrates seamlessly into existing Transformer architectures to address distributional shifts during FM pretraining.
    
    \item Through comprehensive experiments across diverse time series tasks, including classification and forecasting, we demonstrate substantial improvements in both fine-tuning performance and generalization capabilities compared to conventional approaches, validating the efficacy of our distribution-aware learning paradigm.
\end{itemize}

\section{Related Work}
\subsection{Foundation Models for Time Series}

Foundation models (FMs) have gained attention in TS analysis, following the success of Large Language Models (LLMs) in natural language processing (NLP) \cite{FM_ts_survey}. However, while some studies have adapted pretrained LLMs for TS data \cite{LLM_TEMPO, FM_ts_Lag_Llama, FM_ts_UniTS, FM_ts_OneFitAll}, this approach is not ideal for TS tasks. The inherent differences between text, which is discrete and categorical, and TS data, which is continuous and numeric, present significant challenges for LLM-based methods \cite{FM_ts_UniCL}. These models often fail to capture the unique temporal patterns and dynamics of TS data.
Other research has focused on designing FMs specifically for TS tasks \cite{FM_ts_TimesFM, FM_ts_Timer, FM_ts_TimeSiam}, often using self-supervised learning techniques like masked sequence prediction \cite{FM_ts_Moment, FM_ts_TiMAE}, contrastive learning \cite{ca_tcc, FM_ts_TimeCLR}, or hybrid methods \cite{FM_ts_LearnEmbed, FM_ts_Simmtm}. However, it's vital to distinguish works based on their pretraining strategy. Some methods train on a single dataset and test on that same dataset, such as PatchTST \cite{FM_ts_PatchTST} and TSLANet \cite{FM_ts_TSLANet}. While these approaches can achieve strong performance within a specific domain, they do not involve pretraining on multiple datasets, limiting their ability to generalize across diverse TS domains.
On the other hand, certain methods adopt a more generalizable approach by pretraining on a pool of datasets \cite{FM_ts_UniCL, FM_ts_Moirai, FM_ts_Chronos}, aiming to build foundation models that can generalize well. 
However, even among these models, some fail to fully address the challenges posed by distribution shifts during pretraining, which can impact their efficacy in real-world applications across different domains.

\subsection{Distribution Shifts in Time Series}

Time series data is particularly prone to distribution shifts due to factors such as changes in sensor behavior, environmental variations, and temporal dynamics \cite{TS_distribution_shift}. A growing body of research aims to mitigate these shifts in deep learning models through techniques such as domain adaptation \cite{UDA_TS1, UDA_TS2, UDA_TS3, UDA_TS4} and domain generalization \cite{DG_TS1, DG_TS3}. These approaches seek to capture domain-invariant features that can be generalized across different distributions. Besides, architecture-specific mechanisms have been developed, including Adaptive RNNs \cite{AdaRNN}, Non-stationary Transformers \cite{NonsTransformer}, Instance Normalization flows \cite{DishTS, INFlows}, and contextualized adapters \cite{SOLID}. These mechanisms aim to alleviate the impact of non-stationary factors through distribution characterization. However, a significant drawback of these designs is their limited transferability across different model architectures, potentially hindering their broader applicability in diverse TS analysis scenarios.
Beyond architecture-specific designs, several normalization-based strategies have been proposed to address distribution shifts in TS data \cite{AdaNorm, DAIN}. For instance, RevIN \cite{RevIN} introduced instance normalization to mitigate distribution shifts by leveraging statistics from individual samples to normalize TS data. Despite these advances, the application of such techniques to Transformer architectures remains limited, and their utilization in multi-dataset training scenarios is still underexplored.

\subsection{Adaptive Normalization Techniques}

Adaptive normalization methods, in contrast to traditional fixed schemes, learn flexible strategies to address covariate shift \cite{Survey_AdaNorm, ASR_Norm}. For instance, Adaptive Batch Normalization dynamically adjusts normalization parameters across batches, while Adaptive Instance Normalization aligns channel-wise mean and variance to match style input \cite{AdaBN, dsbn, BeyondBN}.
Recent research has focused on developing adaptive normalization techniques specifically for the non-stationary characteristics of TS data \cite{ST_Norm, AdaNorm}. For example, DAIN introduced a non-linear network for adaptive input normalization \cite{DAIN}, which was subsequently extended by various approaches \cite{BIN, ExtendDAIN}. These extensions incorporated adaptive preprocessing layers into deep neural networks. RevIN proposed a symmetric, model-agnostic method that normalizes input sequences and denormalizes model output sequences in TS forecasting \cite{RevIN}. More recently, SAN introduced slice-level adaptive normalization, offering more flexible normalization and denormalization for TS forecasting \cite{SAN}, while SIN proposed selective and interpretable normalization to select statistics and learn the normalization transformation \cite{SIN}.
While existing normalization methods have shown efficacy, they assume uniform statistical properties across all TS instances, which may not be optimal while pretraining with multiple datasets. In contrast, we explicitly take the distribution inconsistencies into consideration during FM pretraining, offering a more nuanced and effective training strategy.

\begin{figure*}[!thb]
\begin{center}
\includegraphics[width=0.9 \textwidth]{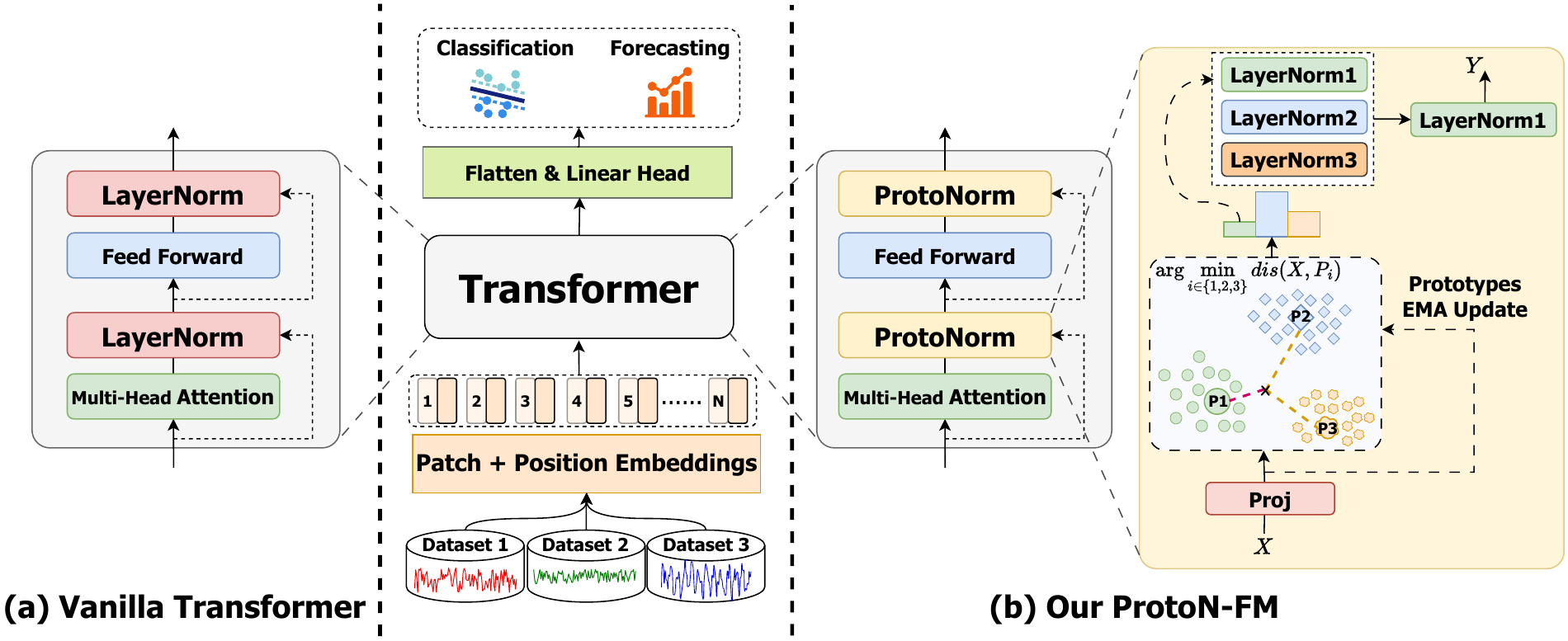}
\end{center}
\caption{Framework comparison between vanilla Transformer and our \abb. (a) Vanilla Transformer with standard LayerNorm, employing fixed normalization parameters across all inputs. (b) Our \abb with \layer mechanism for dynamic LayerNorm assignment via prototype-guided gating. Each input is assigned appropriate LayerNorm parameters based on its similarity to learned prototypes, where these prototypes undergo continuous refinement through EMA updates during training.}
\label{fig:framework}
\end{figure*}

\section{Proposed Method}

\subsection{Preliminaries}
\subsubsection{Problem Definition}
Given a collection of time series datasets $\mathcal{D} = \{\mathcal{D}_k | k = 1, 2, ..., n\}$, each dataset $\mathcal{D}_k$ encompasses a variable number of samples with dimensions $L_k \times C_k$ ($L_k$ denoting signal length and $C_k$ representing the number of sensors or variables). Our objective is to pretrain a time series foundation model $\mathcal{M}$ on this collection $\mathcal{D}$ while addressing inter-dataset distributional shifts. The model subsequently undergoes fine-tuning on either a novel or known dataset utilizing limited data samples to achieve superior performance.

\subsubsection{Layer Normalization}
Layer Normalization (LN) \cite{layernormalization} constitutes a fundamental training mechanism in deep learning networks, particularly in the prevalent Transformer architecture \cite{transformer}. Unlike Batch Normalization (BN), which normalizes across the batch dimension, LN normalizes across features within a single layer. Analogous to BN, LN also incorporates two trainable affine parameters $\gamma$ and $\beta$ to enable network learning of distinct scales and shifts. Given a layer's activation $x \in \mathbb{R}^{C \times L}$ for a single input, LN is formulated as,
\begin{equation}
    LN \left(x_i; \gamma, \beta\right) = \gamma \cdot \hat{x}_i + \beta,
\end{equation}
where 
\begin{equation}
    \hat{x}_i = \frac{x_i-\mu}{\sqrt{\sigma^2+\epsilon}}.
\end{equation}
Here, $\mu$ denotes the mean and $\sigma^2$ represents the variance computed over layer features for a single input,
\begin{equation}
    \mu = \frac{1}{d}\sum_{i=1}^{d}x_i, \quad \sigma^2 = \frac{1}{d}\sum_{i=1}^{d}{(x_i - \mu)}^2,
\end{equation}
and $\epsilon$ is a small constant to prevent divide-by-zero.

During training, LN normalizes activations by computing mean and variance across features for each sample independently, mitigating internal covariate shift and enhancing training stability. During inference, LN executes identical normalization operations but with frozen parameters, implementing the learned scale and shift transformations. Unlike BN, LN's sample-wise normalization eschews running statistics, ensuring consistent behavior between training and inference.

\subsection{Prototype-guided Dynamic Normalization Mechanism}
It is important to explain why we chose to modify LN specifically, rather than other components of the Transformer, to address the distribution shift. LN is an ideal candidate for this modification because it has fewer parameters than other parts of the Transformer, making it computationally efficient to replicate. This allows us to handle variations across different datasets while minimizing the risk of overfitting.

Moreover, previous empirical evidence demonstrates that domain-specific normalization approaches, such as BatchNorm, excel in mitigating domain shifts during adaptation tasks \cite{dsbn}. This insight motivated our development of a domain-aware normalization framework tailored for time series data. However, conventional LN approaches presume static relationship between input samples and their corresponding normalization strategies, potentially constraining the model's adaptability to both intra- and inter-dataset variations. Relying on a fixed normalization strategy across entire datasets may inadequately address challenges such as heterogeneous sample characteristics or cross-domain convergence.

\paragraph{Prototype-Guided Gating Network} To overcome these limitations, we introduce \abb, which implements an adaptive and dynamic normalization mechanism, as illustrated in Figure~\ref{fig:framework}. Departing from fixed LN implementations per dataset, we propose \layer layer, comprising multiple LN modules, where selection occurs through a prototype-guided gating network that assigns each sample to its optimal LN based on prototype proximity. Post-pretraining, these learned prototypes function as distributional anchors, enabling sample-specific normalization strategy adaptation.

\begin{figure}[t]
\centering
\includegraphics[width=0.75 \linewidth]{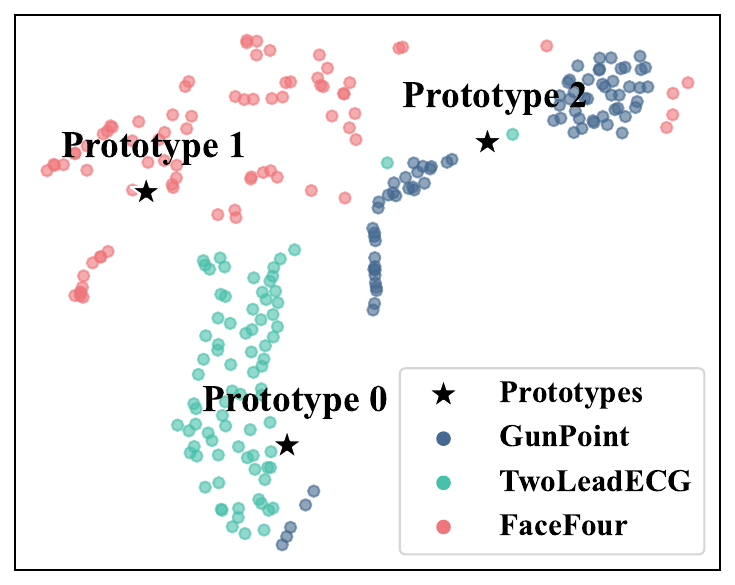}
\caption{Visualization of learned prototypes and sample features. Prototypes capture the unique distribution patterns of each cluster.}
\label{fig:prototypes_distribution}
\end{figure}

Figure~\ref{fig:prototypes_distribution} illustrates how learned prototypes function as distributional anchors, capturing distinct data distributions. This framework enables optimal normalization strategy selection per sample, accommodating both intra- and inter-dataset variations while enhancing the model's capacity to handle complex distributional overlaps.

Formally, each \layer layer incorporates $n$ LayerNorm modules $\{{LN}_1, {LN}_2, ..., {LN}_n\}$, alongside a prototypes-guided gating network $\mathcal{G}$. Considering a signal $v$ with features $x$, $\mathcal{G}$ determines the appropriate LayerNorm for normalization by computing distances between $x$ and predefined prototypes $\{ p_1, p_2, \dots, p_n \}$, where each prototype is associated with a specific LayerNorm module.

\paragraph{Adaptive Normalization} The network selects LayerNorm module ${LN}_i$ whose prototype $p_i$ minimizes the distance to $x$,  implementing optimal normalization mapping. This selection is given by:
\begin{equation}
    i^* = \arg \min_{i \in \{1, 2, \dots, n\}} d(x, p_i),
\end{equation}
where $d(x, p_i)$ represents the distance metric (e.g., Euclidean distance) between $x$ and prototype $p_i$.

\paragraph{Prototype Update} The prototypes evolve during training via Exponential Moving Average (EMA) \cite{Adam}, ensuring adaptive refinement based on distributional dynamics. Formally, the prototype $p_i$ is updated as:
\begin{equation}
    p_i^{(t+1)} = (1- \alpha) \cdot p_i^{(t)} + \alpha \cdot x,
\end{equation}
where $p_i^{(t)}$ denotes the prototype at time $t$, $x$ represents the current cluster's input features, and $\alpha$ constitutes the EMA decay factor. This mechanism ensures continuous prototype refinement while maintaining distributional robustness.

\paragraph{Orthogonality Constraint} To maintain prototype distinctiveness, we incorporate an additional orthogonality constraint. The prototypes initialize with orthogonal parameters, facilitating enhanced discriminative capacity among diverse input features and distributions. Additionally, inspired by \cite{insight_weight_reg}, we implement regularization to preserve prototype independence. Formally, given a matrix $P \in \mathbb{R}^{n \times d}$ where each row represents a prototype, the orthogonal loss is defined as:
\begin{equation}
    \mathcal{L}_{\text{orth}} = \| P P^T - I \|_F^2,
\end{equation}
where $I$ is identity matrix, and $ \| \cdot \|_F^2 $ is Frobenius norm.

\subsection{Self-supervised Pretraining Framework}
Our pretraining framework employs augmentation-based contrastive learning for robust temporal representation acquisition. For input TS sample $x$, we apply two augmentation techniques: time-shift and scaling with jitter \cite{ca_tcc}, generating two diverse views of the same sample, denoted as $\widetilde{x}_1$ and $\widetilde{x}_2$. Time-shift augmentation introduces variations in signal timing by shifting the input sequence along the temporal axis, while scaling with jitter applies random scaling factors combined with small perturbations, simulating variability in signal amplitude and sensor noise.

The encoder and projector head transform $\widetilde{x}_1$ and $\widetilde{x}_2$ into representations $z_1$ and $z_2$. We then utilize NT-Xent loss \cite{simclr} to optimize view similarity while maintaining sample distinctiveness. For N-sample batches, the NT-Xent loss follows:
\begin{equation}
    \mathcal{L}_{\text{NT-Xent}} = - \log \frac{\text{exp}(\text{sim}(z_1, z_2)/\tau)}{\sum_{j=1}^{2N}\mathbf{1}_{[j \neq i]} \text{exp}(\text{sim}(z_i, z_j)/ \tau)},
\end{equation}
where $\text{sim}(z_1, z_2) = \frac{z_1 \cdot z_2}{\|z_1\| \|z_2\|}$ denotes the dot product between $\ell_2$ normalized $z_1$ and $z_2$ (i.e., cosine similarity), $\tau$ represents temperature scaling, and $\mathbf{1}_{[j \neq i]}$ is an indicator function excluding the positive pair from the denominator.

The comprehensive pretraining objective integrates contrastive and orthogonal losses, ensuring robust representation learning and distinct, separable prototypes:
\begin{equation}
    \mathcal{L} = \mathcal{L}_{\text{NT-Xent}} + \lambda \cdot \mathcal{L}_{\text{orth}},
\end{equation}
where $\lambda$ is a hyperparameter that balances the contribution of the orthogonal loss, empirically set to $0.001$ in our experiments.

\subsection{Modular Design for Integration with Existing Models} 
Our \layer mechanism is designed as a modular component that can directly replace standard LN layers in existing Transformer-based architectures. This enables us to evaluate the impact of our normalization approach both as part of our complete pretraining framework and as a drop-in replacement in state-of-the-art models like MOMENT \cite{FM_ts_Moment} and Moirai \cite{FM_ts_Moirai}, without altering their original training objectives or architectures.

\section{Experiments}
We conduct comprehensive evaluations to assess the efficacy of \abb across time series classification and forecasting tasks. The detailed experimental results and comprehensive model analyses are provided in subsections \ref{sec:appendix:classification_detailed_results}, \ref{sec:appendix:forecasting_detailed_results}, and \ref{sec:appendix:model_analysis} in the Appendix.

\begin{table*}[!t]
\centering
\caption{Classification results in different datasets. Results are averaged across each subset of datasets. We calculate the Accuracy and F1-score (\%) for each dataset. \textcolor{blue}{\textbf{Blue}}: best results, \textcolor{purple}{\underline{Purple}}: second best. Full results are listed in Tables \ref{tab:ucr_detailed}, \ref{table:phm_indomain}, and \ref{table:har_indomain} in Appendix.}
{\fontsize{9}{11} \selectfont
\setlength{\tabcolsep}{1.0mm}{
\renewcommand\arraystretch{1.1}
\begin{tabular}{@{}c|cccc|cccc@{}} 
\toprule 
\multirow{2}*{Datasets} & \multicolumn{4}{c|}{Accuracy} & \multicolumn{4}{c}{Macro-F1} \\ \cline{2-9}
 & Sup. & Individual & Vanilla & \abb & Sup. & Individual & Vanilla & \abb \\ \midrule
UCR Archive (91 datasets)  & 62.03 & 61.53 & \textcolor{purple}{\underline{66.66}} & \textcolor{blue}{\textbf{67.78}} & 48.30 & 47.65 & \textcolor{purple}{\underline{52.50}} & \textcolor{blue}{\textbf{53.51}} \\
Machine Fault Diagnosis (3 datasets)  & 50.58 & 56.17 & \textcolor{purple}{\underline{66.30}} & \textcolor{blue}{\textbf{70.33}} & 46.88 & 54.03 & \textcolor{purple}{\underline{61.95}} & \textcolor{blue}{\textbf{67.13}} \\
Human Activity Recognition (5 datasets) & 45.29 & 47.43 & \textcolor{purple}{\underline{48.83}} & \textcolor{blue}{\textbf{51.05}} & 34.48 & 36.25 & \textcolor{purple}{\underline{37.15}} & \textcolor{blue}{\textbf{38.90}} \\ \hline
Average & 52.63 & 55.04 & \textcolor{purple}{\underline{60.60}} & \textcolor{blue}{\textbf{63.05}} & 43.22 & 45.98 & \textcolor{purple}{\underline{50.53}} & \textcolor{blue}{\textbf{53.18}} \\
\bottomrule
\end{tabular}
}}
\label{table:avg_classification}
\end{table*}

\subsection{Experimental Setup}
\label{sec:appendix:experimental_setup}

\paragraph{Evaluation Strategy} 
Our validation framework employs a dual-pronged approach across time series domains. For classification, we evaluate the complete \abb framework against established baselines with fine-tuning and assess \layer's specific contribution through MOMENT \cite{FM_ts_Moment} integration in zero-shot settings. For forecasting, we evaluate \layer's effectiveness by incorporating it into Moirai \cite{FM_ts_Moirai} across both in-distribution and out-of-distribution scenarios. This approach isolates our normalization mechanism's impact while demonstrating its broader applicability as a modular enhancement to foundation models.

\paragraph{Model Architecture and Integration}
We adopt PatchTST \cite{FM_ts_PatchTST} as our base architecture for its balanced performance and efficiency. The encoder adheres to standard Transformer design principles: patches undergo embedding transformation before processing through multiple layers of multi-head attention with \layer and feed-forward networks with subsequent \layer (Figure~\ref{fig:framework}).
Besides, to validate the effectiveness of our \layer, we integrate it into two recent representative foundation models—MOMENT \cite{FM_ts_Moment} for classification and Moirai \cite{FM_ts_Moirai} for forecasting—while maintaining their original architectures and training protocols, only replacing their normalization layers with our prototype-guided mechanism. For computational efficiency and fair comparison, we utilize the small variants of both models (MOMENT$_{small}$ and Moirai$_{small}$) in our experiments. 

\paragraph{Handling Varying Time Series Characteristics}
To address dataset heterogeneity, we implement standardized preprocessing across tasks. For variations in channel count, we employ channel replication from samples with fewer channels to achieve parity with the maximum channel count in the pretraining dataset ensemble, incorporating stochastic noise injection to mitigate overfitting risks. Regarding sequence length disparities, we utilize adaptive downsampling for extended sequences and zero-padding for truncated ones, standardizing to prescribed temporal dimensions: 512 timesteps for UCR datasets, 1024 for MFD tasks, and 128 for HAR tasks. For MOMENT and Moirai baselines, we maintain their original preprocessing settings to ensure fair comparison. These techniques ensure consistent input dimensions while preserving data characteristics.

\paragraph{Hyperparameters}
We optimize our model utilizing the AdamW optimizer with a learning rate of $1e-3$, weight decay of $1e-5$, and dropout rate of $0.15$. A cosine learning rate scheduler with $2000$ warmup steps is applied across all tasks. For UCR and MFD tasks, we use a pretraining batch size of $256$ over $5$ epochs, with embedding dimension of $256$, $8$ attention heads, $12$ encoder layers, and patch size of $50$. The input sequence length is set to $512$ for UCR and $1024$ for MFD tasks. Fine-tuning maintains this architecture but adjusts batch sizes to $32$ for UCR and $64$ for MFD tasks, extending training to $50$ epochs.
For HAR tasks, we use a pretraining batch size of $128$ over $5$ epochs, with embedding dimension of $128$, $8$ attention heads, $6$ encoder layers, patch size of $32$, and input sequence length of $128$; fine-tuning reduces the batch size to $8$ over $50$ epochs. When comparing with MOMENT and Moirai baselines, we maintain their original hyperparameter settings. Model performance was evaluated using accuracy and macro-averaged F1 scores as primary metrics. Each experiment was repeated three times, with the average performance reported. Implementation uses PyTorch on NVIDIA L$40$ GPUs.

\paragraph{Data Preprocessing}
We conduct comprehensive empirical evaluations across diverse time series domains, encompassing classification and forecasting tasks. For classification experiments, we leverage the 91 UCR archive datasets \cite{UCR_datasets} with their predetermined train/test splits, further segmenting the training corpus into training (80\%) and validation (20\%) subsets. The model undergoes pretraining on the consolidated UCR collection prior to fine-tuning and evaluation on individual datasets. Notably, for baseline comparisons with MOMENT \cite{FM_ts_Moment}, we maintain methodological consistency with their experimental protocols. For MFD tasks, we augment the pretraining phase by incorporating three supplementary prognostics and health management (PHM) datasets: CWRU \cite{CWRU}, FEMTO \cite{FEMTO}, and XJTUSY \cite{XJTUSY}.  Subsequently, the model undergoes fine-tuning and evaluation on three target datasets (IMS \cite{IMS}, UO \cite{UO}, and PU \cite{PU}). Analogously, for HAR tasks, we utilize five datasets (HHAR \cite{HHAR}, SKODA \cite{SKODA}, UCIHAR \cite{UCIHAR}, USCHAD \cite{USCHAD}, and WISDM \cite{WISDM}) during pretraining, followed by individual dataset evaluation. Both MFD and HAR datasets, lacking predefined partitions, are systematically segmented into train/validation/test sets with a 60/20/20 ratio. For forecasting experiments, we strictly adhere to the experimental methodology of Moirai \cite{FM_ts_Moirai} to ensure fair comparison, encompassing their data preprocessing protocols, training paradigms, and evaluation metrics.

\begin{table*}[!t]
\caption{Forecasting performance comparison with different prediction lengths. Results show the Mean Absolute Error (MAE) and Mean Squared Error (MSE) metrics for Moirai and \abb models across multiple benchmarking datasets with different prediction lengths $\in \{96, 192, 336, 720\}$. \textcolor{blue}{\textbf{Blue}}: best results. The rightmost columns show averaged performance across all prediction lengths.}
\label{tab:forecasting-results}
\centering
{\fontsize{9}{11}\selectfont
\setlength{\tabcolsep}{0.9mm}
\renewcommand\arraystretch{1.1}
\begin{tabular}{lc|cc|cc|cc|cc|cc}
\hline
\multirow{3}{*}{Datasets} & \multicolumn{1}{c}{\multirow{3}{*}{Metric}} & \multicolumn{10}{c}{Prediction Length} \\
\cline{3-12}
& \multicolumn{1}{c}{} & \multicolumn{2}{c}{96} & \multicolumn{2}{c}{192} & \multicolumn{2}{c}{336} & \multicolumn{2}{c}{720} & \multicolumn{2}{c}{Average} \\
\cline{3-12}
& \multicolumn{1}{c}{} & Moirai & \texttt{ProtoN-FM} & Moirai & \texttt{ProtoN-FM} & Moirai & \texttt{ProtoN-FM} & Moirai & \texttt{ProtoN-FM} & Moirai & \texttt{ProtoN-FM} \\
\hline
\multirow{2}{*}{ETTh1} 
& \textbf{MAE} & \textcolor{blue}{\textbf{0.4033}} & 0.4070 & \textcolor{blue}{\textbf{0.4209}} & 0.4251 & \textcolor{blue}{\textbf{0.4314}} & 0.4364 & \textcolor{blue}{\textbf{0.4473}} & 0.4596 & \textcolor{blue}{\textbf{0.4257}} & 0.4320 \\
& \textbf{MSE} & 0.3756 & \textcolor{blue}{\textbf{0.3747}} & \textcolor{blue}{\textbf{0.4000}} & 0.4005 & \textcolor{blue}{\textbf{0.4112}} & 0.4146 & \textcolor{blue}{\textbf{0.4170}} & 0.4332 & \textcolor{blue}{\textbf{0.4010}} & 0.4058 \\
\hline
\multirow{2}{*}{ETTh2} 
& \textbf{MAE} & \textcolor{blue}{\textbf{0.3329}} & 0.3334 & 0.3734 & \textcolor{blue}{\textbf{0.3729}} & 0.3922 & \textcolor{blue}{\textbf{0.3879}} & 0.4110 & \textcolor{blue}{\textbf{0.4105}} & 0.3774 & \textcolor{blue}{\textbf{0.3762}} \\
& \textbf{MSE} & \textcolor{blue}{\textbf{0.2868}} & 0.2884 & \textcolor{blue}{\textbf{0.3487}} & 0.3490 & 0.3709 & \textcolor{blue}{\textbf{0.3667}} & \textcolor{blue}{\textbf{0.3848}} & 0.3860 & 0.3478 & \textcolor{blue}{\textbf{0.3475}} \\
\hline
\multirow{2}{*}{ETTm1} 
& \textbf{MAE} & \textcolor{blue}{\textbf{0.3681}} & 0.3708 & \textcolor{blue}{\textbf{0.3820}} & 0.3850 & \textcolor{blue}{\textbf{0.3962}} & 0.3976 & 0.4187 & \textcolor{blue}{\textbf{0.4167}} & \textcolor{blue}{\textbf{0.3913}} & 0.3925 \\
& \textbf{MSE} & 0.3698 & \textcolor{blue}{\textbf{0.3684}} & \textcolor{blue}{\textbf{0.3785}} & 0.3829 & \textcolor{blue}{\textbf{0.3955}} & 0.4027 & \textcolor{blue}{\textbf{0.4261}} & 0.4291 & \textcolor{blue}{\textbf{0.3925}} & 0.3958 \\
\hline
\multirow{2}{*}{ETTm2} 
& \textbf{MAE} & 0.2869 & \textcolor{blue}{\textbf{0.2781}} & 0.3264 & \textcolor{blue}{\textbf{0.3181}} & 0.3596 & \textcolor{blue}{\textbf{0.3569}} & \textcolor{blue}{\textbf{0.4063}} & 0.4141 & 0.3448 & \textcolor{blue}{\textbf{0.3418}} \\
& \textbf{MSE} & 0.2135 & \textcolor{blue}{\textbf{0.2042}} & 0.2768 & \textcolor{blue}{\textbf{0.2683}} & \textcolor{blue}{\textbf{0.3304}} & 0.3310 & \textcolor{blue}{\textbf{0.4047}} & 0.4262 & \textcolor{blue}{\textbf{0.3064}} & 0.3074 \\
\hline
\multirow{2}{*}{Electricity} 
& \textbf{MAE} & 0.2868 & \textcolor{blue}{\textbf{0.2799}} & 0.2989 & \textcolor{blue}{\textbf{0.2906}} & 0.3112 & \textcolor{blue}{\textbf{0.3027}} & 0.3362 & \textcolor{blue}{\textbf{0.3288}} & 0.3083 & \textcolor{blue}{\textbf{0.3005}} \\
& \textbf{MSE} & 0.1948 & \textcolor{blue}{\textbf{0.1848}} & 0.2106 & \textcolor{blue}{\textbf{0.1984}} & 0.2247 & \textcolor{blue}{\textbf{0.2132}} & 0.2582 & \textcolor{blue}{\textbf{0.2479}} & 0.2221 & \textcolor{blue}{\textbf{0.2111}} \\
\hline
\multirow{2}{*}{Weather} 
& \textbf{MAE} & 0.2126 & \textcolor{blue}{\textbf{0.2102}} & 0.2493 & \textcolor{blue}{\textbf{0.2460}} & 0.2817 & \textcolor{blue}{\textbf{0.2784}} & 0.3210 & \textcolor{blue}{\textbf{0.3201}} & 0.2662 & \textcolor{blue}{\textbf{0.2637}} \\
& \textbf{MSE} & 0.1760 & \textcolor{blue}{\textbf{0.1729}} & 0.2172 & \textcolor{blue}{\textbf{0.2118}} & 0.2626 & \textcolor{blue}{\textbf{0.2560}} & 0.3189 & \textcolor{blue}{\textbf{0.3171}} & 0.2437 & \textcolor{blue}{\textbf{0.2395}} \\
\hline
\end{tabular}}
\end{table*}


\subsection{Classification Experiments}
\paragraph{Datasets}
We examine the classification efficacy of \abb across 99 diverse datasets, encompassing 91 UCR Archive datasets \cite{UCR_datasets}, 3 machine fault diagnosis (MFD) datasets, and 5 human activity recognition (HAR) datasets. In particular, the MFD datasets include IMS \cite{IMS}, UO \cite{UO}, and PU \cite{PU}, while the HAR datasets consist of HHAR \cite{HHAR}, SKODA \cite{SKODA}, UCIHAR \cite{UCIHAR}, USCHAD \cite{USCHAD}, and WISDM \cite{WISDM}. These datasets exhibit distinct characteristics and span a comprehensive range of time series applications. Detailed dataset descriptions are provided in supplementary material.

\paragraph{Baselines and Experimental Settings}
We evaluate our method using PatchTST architecture \cite{FM_ts_PatchTST} in both with and without fine-tuning settings. For evaluation with fine-tuning, we compare against supervised training (\textit{Sup.}), individual dataset pretraining (\textit{Individual}), and conventional multi-dataset pretraining (\textit{Vanilla}). For MFD and HAR tasks, each utilizes 100 randomly sampled instances for fine-tuning, maintaining a minimum of 5 samples per class when required. For UCR Archive, we employ the complete training samples due to their constrained sample sizes and substantial variation in sample quantities across datasets. During fine-tuning, we substitute the self-supervised head with a linear classifier while maintaining frozen prototypes. For evaluation without fine-tuning, we benchmark against MOMENT \cite{FM_ts_Moment} baseline by incorporating \layer into their framework. All evaluations employ held-out test sets.



\paragraph{Results with Fine-tuning}
Table \ref{table:avg_classification} demonstrates the consistent superiority of our \abb method in time series classification. Multi-dataset pretraining approaches exhibit substantial advantages, with vanilla pretraining attaining 60.60\% average accuracy compared to 55.04\% for individual pretraining and 52.63\% for supervised learning. \abb further elevates performance across all dataset categories. On UCR Archive, \abb attains 67.78\% accuracy and 53.51\% Macro-F1, surpassing vanilla pretraining by 1.12\% and 1.01\% respectively. The improvements are more pronounced in MFD tasks (accuracy: 70.33\% vs. 66.30\%; Macro-F1: 67.13\% vs. 61.95\%) and remain consistent in HAR (51.05\% accuracy, 38.90\% Macro-F1). These empirical findings demonstrate that our prototype-guided normalization effectively mitigates distribution shifts while preserving robust generalization, achieving comprehensive averages of 63.05\% accuracy and 53.18\% Macro-F1 across diverse time series tasks.

\begin{figure}[t]
   \centering
   \subfigure[]{\includegraphics[width=0.48\linewidth]{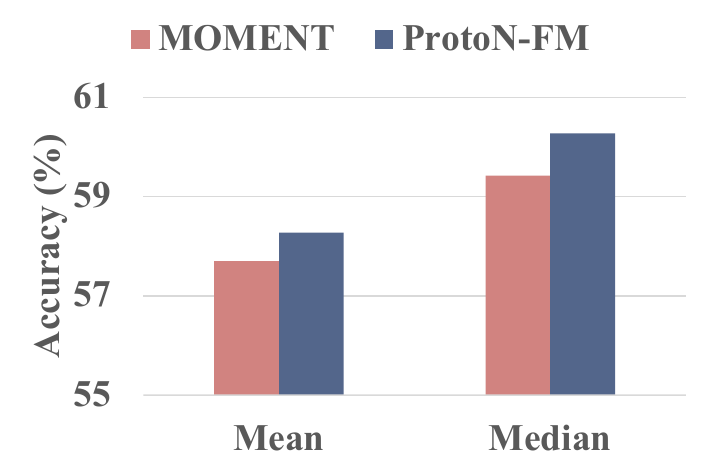}
   \label{fig:moment_without_finetuning}}
   \subfigure[]{\includegraphics[width=0.48\linewidth]{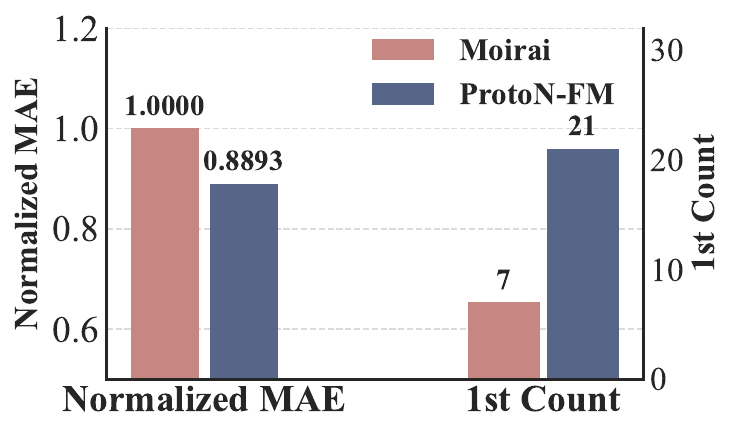}
   \label{fig:normalizedMAE}}
   \caption{Comparative analysis of model performance across classification and forecasting tasks. Full results are listed in Tables \ref{tab:moment_ucr_comparison} and \ref{tab:moirai_iid_mae_comparison} in Appendix. (a) Classification accuracy evaluation across 91 UCR datasets without fine-tuning. (b) Quantitative assessment via normalized MAE metrics and frequency of optimal performance on the Monash benchmark.}
   \label{fig:performance_comparison}
\end{figure}

\paragraph{Results without Fine-tuning}
Figure \ref{fig:moment_without_finetuning} illustrates \abb's efficacy in classification without fine-tuning on UCR Archive. Employing an SVM classifier trained on learned representations, \abb achieves 58.27\% mean accuracy and 60.27\% median accuracy, exceeding MOMENT baseline (57.70\% mean, 59.42\% median) by 0.57\% and 0.85\% respectively. This consistent improvement across both metrics indicates that our prototype-guided normalization mechanism facilitates more robust and transferable representations, even without task-specific adaptation. The superior performance in this challenging paradigm further substantiates our approach's capability to effectively mitigate distribution shifts and extract generalizable patterns from time series data, rendering it particularly valuable in scenarios where labeled data is scarce or unavailable.

\subsection{Forecasting Experiments}
\paragraph{Datasets}
To assess the efficacy of \abb in forecasting tasks, we conduct extensive evaluations across 34 datasets, encompassing 28 datasets from the Monash Time Series Forecasting Archive \cite{Monash_datasets}, and 6 datasets from a subset of the established long sequence forecasting benchmark \cite{LSF, AutoFormer}, comprising four ETT datasets (ETTh1, ETTh2, ETTm1, ETTm2), Electricity, and Weather dataset. Detailed dataset descriptions are provided in supplementary material.

\paragraph{Baselines and Experimental Settings}
We evaluate our \layer through integration with Moirai \cite{FM_ts_Moirai} foundation model across both in-distribution (IID) and out-of-distribution (OOD) scenarios. For IID evaluation, we benchmark on Monash forecasting archive \cite{Monash_datasets} utilizing normalized Mean Absolute Error (MAE) for direct comparison with Moirai baseline. For OOD evaluation, we examine model robustness on unseen datasets from a long sequence forecasting benchmark, employing both MSE and MAE metrics to quantify generalization capabilities across distribution shifts.


\paragraph{Results on In-distribution Forecasting}
Figure \ref{fig:normalizedMAE} illustrates \abb's efficacy on the Monash Time Series Forecasting Benchmark. Our method achieves a normalized MAE of 0.8893, demonstrating an 11.07\% enhancement over the Moirai baseline. Notably, \abb exhibits superior performance at the individual dataset level, attaining optimal results on 21 datasets compared to Moirai's 7 datasets. This substantial performance differential (21 vs. 7) indicates that our prototype-guided normalization facilitates more robust and consistent forecasting capabilities across diverse time series distributions. These empirical findings substantiate our approach's efficacy in elevating foundation model performance for in-distribution forecasting tasks.



\paragraph{Results on Out-of-distribution Forecasting}
Table \ref{tab:forecasting-results} delineates OOD forecasting results across varying prediction lengths and datasets, demonstrating \abb's effectiveness. Our method exhibits particularly robust performance on the Electricity and Weather datasets, consistently surpassing Moirai across all prediction lengths in both metrics. Specifically, on the Electricity dataset, \abb achieves systematic improvements with average MAE reduced from 0.3083 to 0.3005 and MSE from 0.2221 to 0.2111. The performance advantage persists across diverse prediction horizons, indicating robust long-term forecasting capabilities.
For the ETT series datasets, the results exhibit mixed but generally competitive performance. While Moirai demonstrates marginal superiority on ETTh1 and ETTm1, \abb exhibits enhanced performance on ETTh2 and ETTm2. Notably, on ETTm2, our method achieves substantial improvements for shorter prediction lengths (96 and 192 steps), with MAE reduced from 0.2869 to 0.2781 and 0.3264 to 0.3181 respectively. This suggests our prototype-guided normalization mechanism excels at capturing local temporal patterns in specific data typologies. Furthermore, as prediction horizons increase, both models exhibit anticipated performance degradation, yet \abb maintains its competitive advantage or demonstrates more graceful degradation, underscoring its robustness in challenging long-term forecasting scenarios.

\section{Model Analysis}
\begin{table}[t]
\centering
\caption{ Ablation study results on MFD datasets showing average accuracy and macro-F1 scores (\%). \textcolor{blue}{\textbf{Blue}}: best.}
{\fontsize{9}{11}\selectfont
\setlength{\tabcolsep}{0.7mm}{
\renewcommand\arraystretch{1.0}
\begin{NiceTabular}{@{}l|cccc|cccc@{}} 
\toprule 
\multirow{2}*{Variants} & \multicolumn{4}{c|}{Accuracy} & \multicolumn{4}{c}{Macro-F1} \\ \cline{2-9}
 & IMS & UO & PU & \cellcolor{gray!20}Avg & IMS & UO & PU & \cellcolor{gray!20}Avg \\ \midrule
w/o ProtoGate  & 77.51 & 60.43 & 62.02  & \cellcolor{gray!20}66.65 & 69.26 & 59.37 & 58.81 &  \cellcolor{gray!20}62.48   \\
w/o Ortho  & 77.53 & 66.99 & 63.80 &  \cellcolor{gray!20} 69.44 & 70.51 & 66.22  & 60.69 & \cellcolor{gray!20} 65.81 \\ \midrule
\abb  &  78.78 &  68.56 & 63.65  & \cellcolor{gray!20} \textcolor{blue}{\textbf{70.33}} & 73.03 & 67.93 &  60.43 &  \cellcolor{gray!20} \textcolor{blue}{\textbf{67.13}}   \\
\bottomrule
\end{NiceTabular}
}}
\label{table:ablation}
\end{table}
\subsection{Ablation Study}
We conduct systematic ablation experiments to examine the contribution of each critical component within our model. Table \ref{table:ablation} presents comparative analyses across MFD datasets, evaluating \abb against two variants: {w/o ProtoGate} and {w/o Ortho}. The {w/o ProtoGate} variant implements domain-specific LayerNorm, substituting our prototype-guided gate network with dataset-specific LayerNorm selection (elaborated in the supplementary material). The {w/o Ortho} variant eliminates the orthogonality constraints from the prototype learning process.
The empirical results substantiate the efficacy of our design choices. Removing the prototype-guided gate network ({w/o ProtoGate}) yields a substantial performance degradation, with average accuracy declining from 70.33\% to 66.65\% and Macro-F1 from 67.13\% to 62.48\%. This marked deterioration underscores the significance of dynamic distribution matching over static dataset-specific normalization. Similarly, the elimination of orthogonality constraints ({w/o Ortho}) results in diminished performance (69.44\% accuracy, 65.81\% Macro-F1), validating our hypothesis that maintaining prototype diversity through orthogonality constraints facilitates enhanced feature distribution differentiation.

\begin{figure}[t]
    \centering
    \subfigure[]{\includegraphics[width=0.48\linewidth]{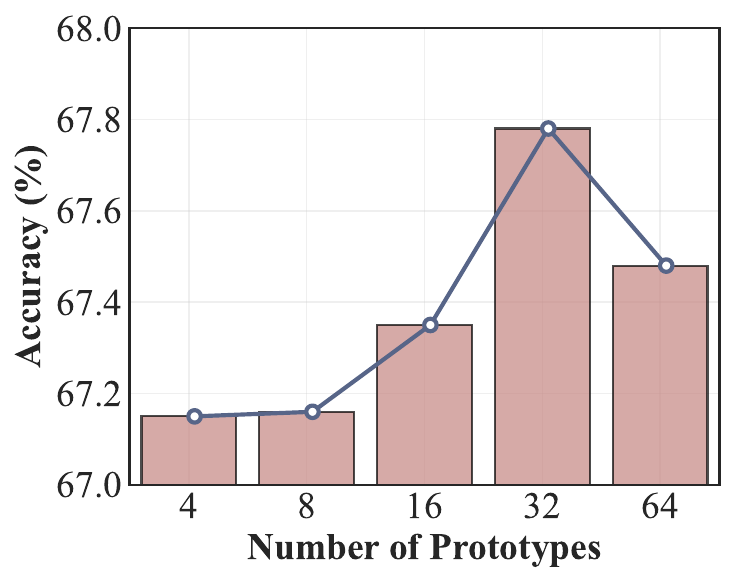}
    \label{figa:prototype_impact}}
    \subfigure[]{\includegraphics[width=0.48\linewidth]{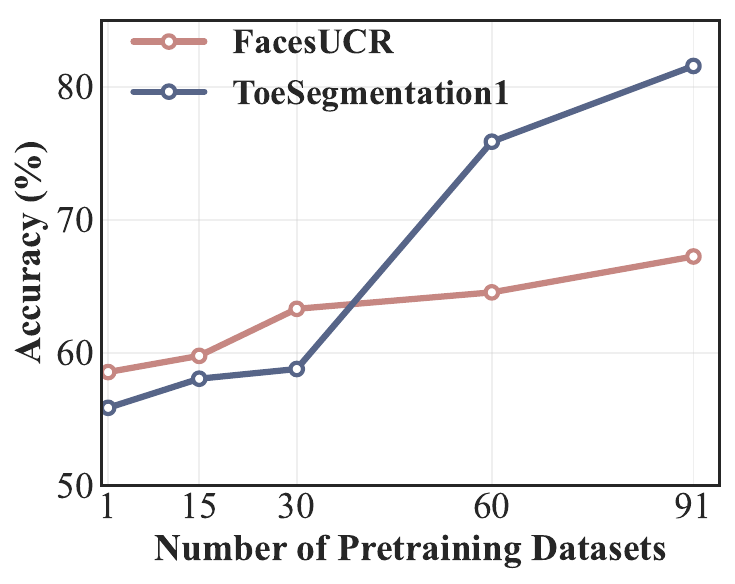}
    \label{figb:pretraining_impact}}
    \caption{Scaling efficiency analysis. (a) Average classification accuracy of \abb across 91 UCR datasets with varying prototype quantities. Full results are listed in Table \ref{tab:different_prototypes_ucr_results} in Appendix. (b) Pretraining dataset scale's impact on classification accuracy, evaluated across two UCR datasets.}
    \label{fig:scaling_efficiency}
\end{figure}

\subsection{Scaling Efficiency}
We conduct comprehensive experiments to examine the scaling efficiency of \abb through prototype quantity and pretraining dataset dimensionality. Figure \ref{fig:scaling_efficiency}(a) illustrates model performance across varying prototype counts $\{4, 8, 16, 32, 64\}$ on 91 UCR datasets. The empirical results demonstrate performance enhancement with increasing prototypes up to 32, attaining peak accuracy at 67.78\%, followed by marginal degradation at 64 prototypes. This indicates that 32 prototypes establish an optimal balance between model capacity and performance, while further increases may induce potential overfitting.
Figure \ref{fig:scaling_efficiency}(b) illustrates the impact of pretraining dataset size on model efficacy, evaluated on two representative UCR datasets. Both exhibit consistent enhancement patterns with increasing pretraining datasets, albeit with distinct scaling characteristics. FacesUCR manifests steady progression from 58.54\% to 67.24\% accuracy, while ToeSegmentation1 demonstrates pronounced improvements, particularly in later stages, escalating from 55.85\% to 81.58\%. This substantial enhancement validates our model's capacity to effectively leverages expanded pretraining data, substantiating the benefits of large-scale pretraining for time series foundation models.

These scaling analyses reveal that \abb efficacy in capturing diverse temporal patterns with optimized prototype quantity while exhibiting robust scaling capabilities with augmented pretraining data, suggesting its potential for even larger-scale applications.

\subsection{Complexity Analysis}
\begin{figure}
    \centering
    \includegraphics[width=0.8 \linewidth]{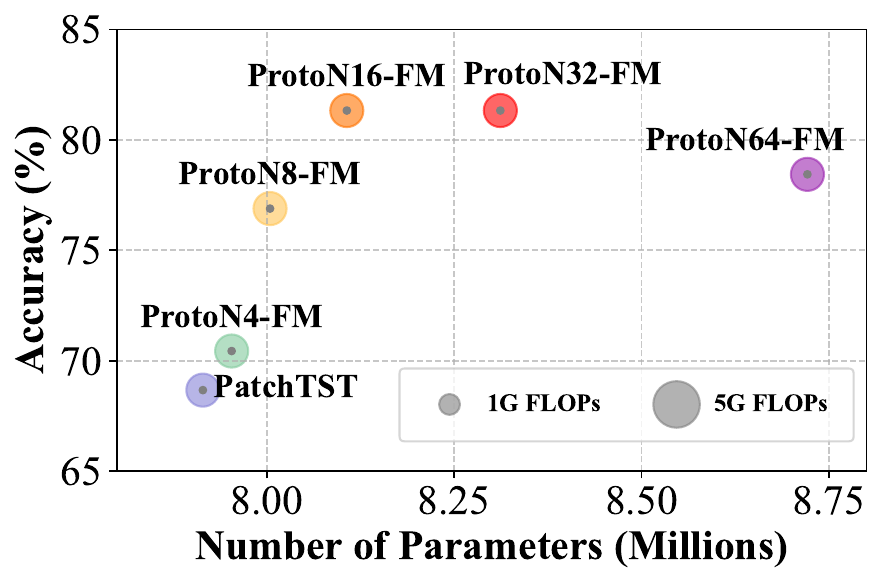}
    \caption{Performance analysis of \abb across varying prototype quantities versus baseline, examining parameter count and FLOPs on UCR GunPoint dataset.}
    \label{fig:flops_comparison}
\end{figure}

We conduct a detailed analysis of our \abb's computational complexity versus baseline PatchTST, as illustrated in Figure \ref{fig:flops_comparison}, examining model parameters, classification accuracy, and computational requirements on the UCR GunPoint dataset.
\abb exhibits minimal parameter overhead: from PatchTST's 7.91M parameters to 7.95M (\texttt{ProtoN4-FM}), 8.00M (\texttt{ProtoN8-FM}), 8.11M (\texttt{ProtoN16-FM}), 8.31M (\texttt{ProtoN32-FM}), and 8.72M (\texttt{ProtoN64-FM}), with maximum increase under 10\%.
Performance-wise, accuracy enhances significantly with increasing prototypes, with \texttt{ProtoN32-FM} attaining optimal accuracy at 81.33\% (versus PatchTST's 68.67\%), representing a substantial 12.66\% improvement. Accuracy demonstrates consistent ascension until 32 prototypes, before marginally declining to 78.44\% with 64 prototypes.
Notably, computational cost maintains 2.53 FLOPs across all prototype configurations, indicating negligible computational overhead from our prototype-guided normalization mechanism. This substantiates \abb's capability to achieve superior performance while preserving computational efficiency, establishing its practicality for real-world deployment.

\subsection{Parameter Analysis on Orthogonal Weights}

\begin{figure}[ht]
    \begin{minipage}{0.49\textwidth}
        \centering
        \includegraphics[width=0.9\linewidth]{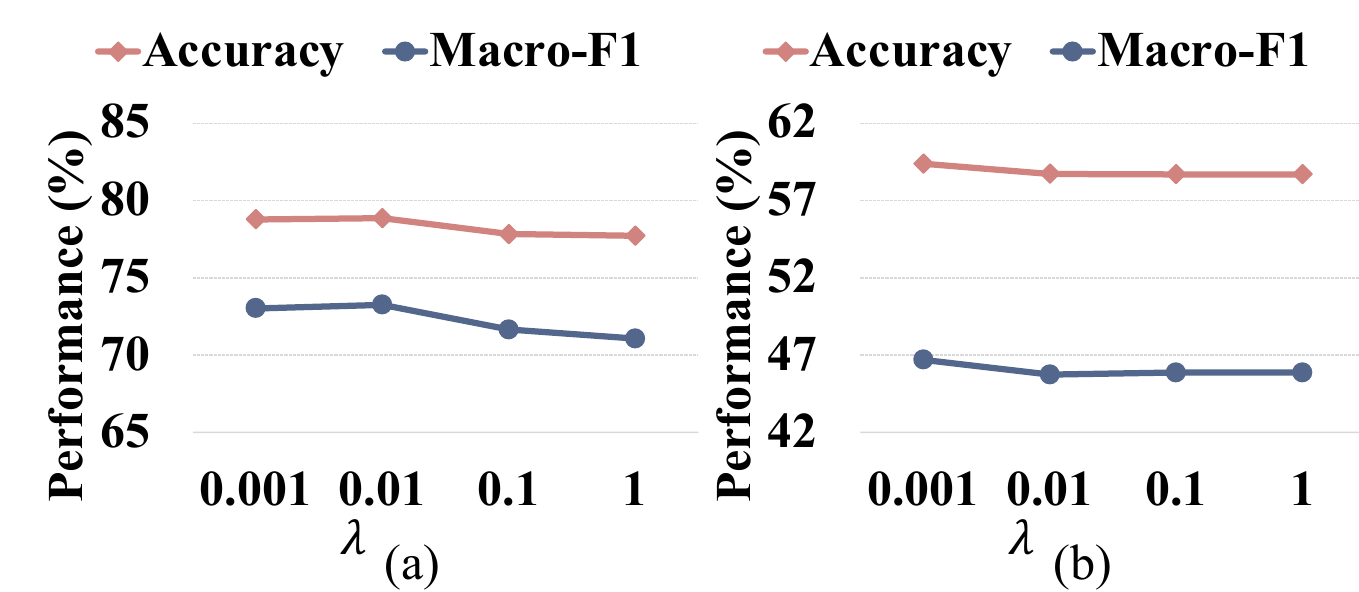}
        \caption{Performance evaluation with varying $\lambda$ parameters on representative datasets. (a) Mean performance metrics on IMS dataset (MFD task). (b) Mean performance metrics on UCIHAR dataset (HAR task).}
        \label{fig:lambda_analysis}
    \end{minipage}
\vspace{0.1cm}
\end{figure}

We conduct additional parameter analyses of orthogonal loss weight to further validate model robustness. These empirical investigations complement our primary analysis through examination of parameter sensitivity across diverse application domains. Figure \ref{fig:lambda_analysis} examines orthogonal loss weight $\lambda \in \{0.001, 0.01, 0.1, 1\}$ impact on IMS (MFD task) and UCIHAR (HAR task) datasets. For IMS, $\lambda = 0.01$ yields optimal efficacy with accuracy of 78.86\% and Macro-F1 score of 73.25\%. Performance degradation at higher $\lambda$ values suggests potential over-regularization. Conversely, the UCIHAR dataset exhibits minimal sensitivity to $\lambda$ variations, with peak accuracy of 59.38\% and Macro-F1 of 46.69\% are observed at $\lambda = 0.001$, maintaining performance stability across varying parameters. These empirical findings indicate robust model performance across $\lambda$ configurations, with smaller values generally sufficient for optimal efficacy.

\begin{figure}[t]
    \centering
    \subfigure[]{\includegraphics[width=0.45\linewidth]{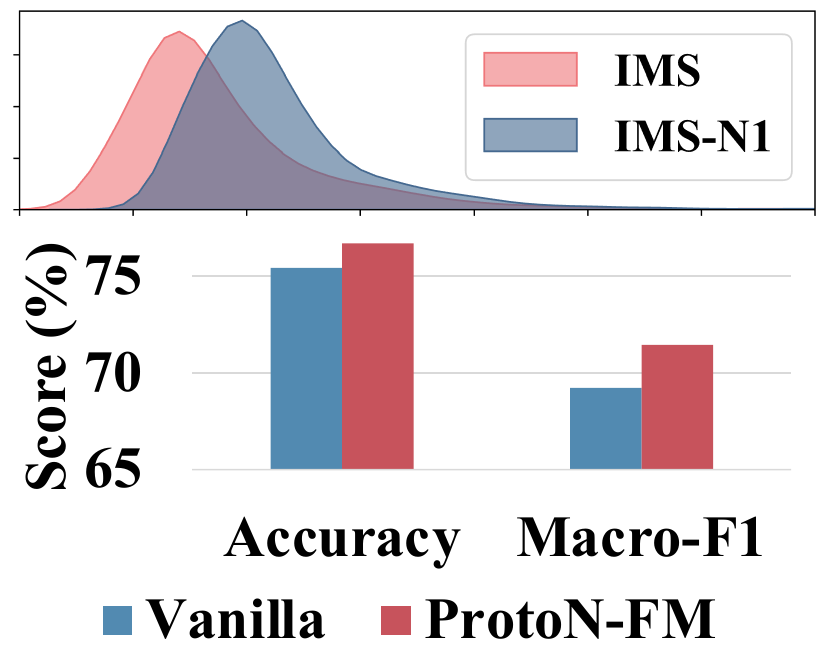}
    \label{figa:distribution_IMS_N1}}
    \subfigure[]{\includegraphics[width=0.45\linewidth]{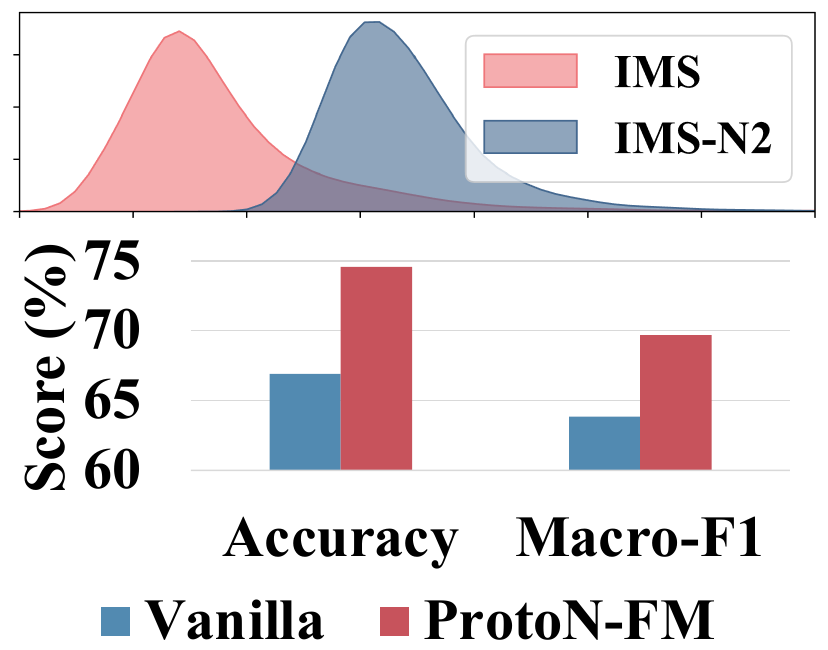}
    \label{figa:distribution_IMS_N2}}\\
    \subfigure[]{\includegraphics[width=0.45\linewidth]{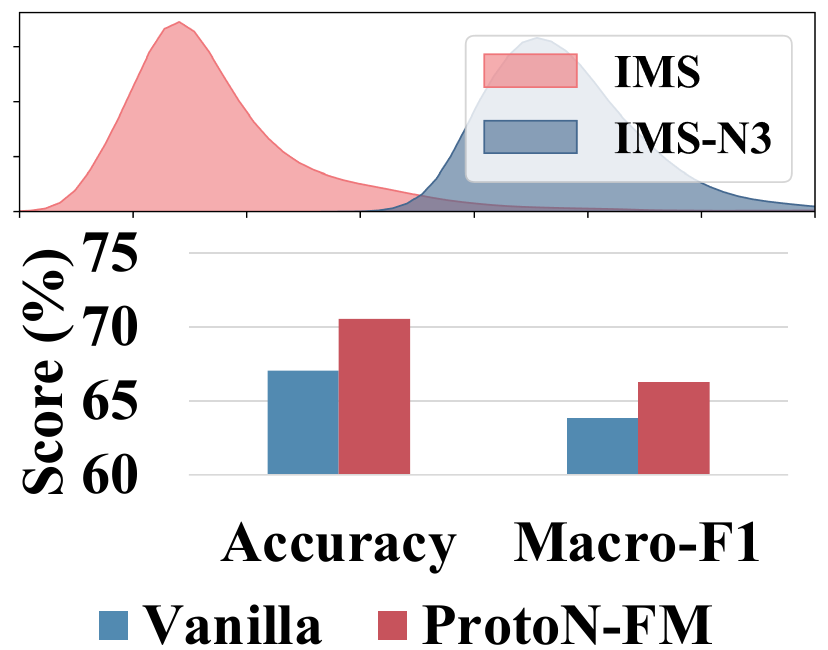}
    \label{figb:distribution_IMS_N3}}
    \caption{Performance analysis under varying distribution shifts. Comparative evaluation of \abb versus vanilla pretraining on IMS, where (a), (b), and (c) represent pretraining scenarios with progressively perturbed IMS variants (IMS-N1, IMS-N2, IMS-N3), followed by fine-tuning and evaluation on original IMS data.}
    \label{fig:distribution_shifts_analysis}
\end{figure}

\subsection{Distribution Shift Analysis}
We examine \abb's robustness under varying levels of distribution shifts utilizing the IMS dataset. We generate three synthetic variants (IMS-N1, IMS-N2, IMS-N3) by introducing Gaussian noise with increasing standard deviations (0.1, 0.2, and 0.3, respectively). The model undergoes pretraining on paired datasets (IMS with IMS-N1, IMS-N2, or IMS-N3) followed by fine-tuning on a constrained IMS subset. Figure~\ref{fig:distribution_shifts_analysis} illustrates comparative performance between \abb and vanilla pretraining approach across perturbation scenarios. The empirical results demonstrate \abb's consistent superiority, particularly in the most challenging scenario (IMS-N3), where it enhances accuracy from 67.05\% to 70.54\% and Macro-F1 from 63.84\% to 66.28\%. These substantial performance gains substantiate our prototype-guided normalization strategy's efficacy in mitigating distribution shifts.

\section{Conclusion}
This paper introduces \abb, a novel approach addressing distributional discrepancies between foundation model pretraining and time series data. \abb facilitates adaptive normalization through prototype-based similarity assessment. Unlike conventional LayerNorm's fixed normalization parameters, our approach learns distinctive prototypes capturing diverse data characteristics, each coupled with a specialized LayerNorm module. Through comprehensive empirical evaluation across time series tasks, we demonstrate \abb's consistent superiority over traditional approaches, particularly under significant distribution shifts. Future research directions encompass exploring universal capabilities across additional downstream tasks and integrating \layer into diverse Transformer architectures to maximize its pretraining potential.

\bibliographystyle{IEEEtran}
\bibliography{main}


\appendix


\begin{table*}[h]
\centering
\caption{Detailed performance comparison of various methods on MFD datasets. We calculate the Accuracy and F1-score (\%) for each dataset. \textcolor{blue}{\textbf{Blue}}: best, \textcolor{purple}{\underline{Purple}}: second best.}
{\fontsize{9}{11}\selectfont
\setlength{\tabcolsep}{2mm}{
\renewcommand\arraystretch{1.0}
\begin{NiceTabular}{@{}l|cccc|cccc@{}} 
\toprule 
\multirow{2}*{Datasets} & \multicolumn{4}{c|}{Accuracy} & \multicolumn{4}{c}{Macro-F1} \\ \cline{2-9}
 & Sup. & Individual & Vanilla & \abb & Sup. & Individual & Vanilla & \abb \\ \midrule
IMS  & 54.22 & 59.48 & \textcolor{purple}{\underline{77.00}} & \textcolor{blue}{\textbf{78.78}} & 47.84 & 57.79 & \textcolor{purple}{\underline{68.39}} & \textcolor{blue}{\textbf{73.03}} \\
UO   & 49.32 & 50.62 & \textcolor{purple}{\underline{60.00}} & \textcolor{blue}{\textbf{68.56}} & 48.20 & 49.33 & \textcolor{purple}{\underline{58.81}} & \textcolor{blue}{\textbf{67.93}} \\
PU   & 48.19 & 58.42 & \textcolor{purple}{\underline{61.91}} & \textcolor{blue}{\textbf{63.65}} & 44.61 & 54.98 & \textcolor{purple}{\underline{58.66}} & \textcolor{blue}{\textbf{60.43}} \\ \midrule \rowcolor{gray!15}
Average & 50.58 & 56.17 & \textcolor{purple}{\underline{66.30}} & \textcolor{blue}{\textbf{70.33}} & 46.88 & 54.03 & \textcolor{purple}{\underline{61.95}} & \textcolor{blue}{\textbf{67.13}} \\
\bottomrule
\end{NiceTabular}
}}
\label{table:phm_indomain}
\end{table*}


\begin{table*}[h]
\centering
\caption{Detailed performance comparison of various methods on HAR datasets. We calculate the Accuracy and F1-score (\%) for each dataset. \textcolor{blue}{\textbf{Blue}}: best, \textcolor{purple}{\underline{Purple}}: second best.}
{\fontsize{9}{11}\selectfont
\setlength{\tabcolsep}{2mm}{
\renewcommand\arraystretch{1.0}
\begin{NiceTabular}{@{}l|cccc|cccc@{}} 
\toprule 
\multirow{2}*{Datasets} & \multicolumn{4}{c|}{Accuracy} & \multicolumn{4}{c}{Macro-F1} \\ \cline{2-9}
 & Sup. & Individual & Vanilla & \abb & Sup. & Individual & Vanilla & \abb \\ \midrule
HHAR   & 69.57 & 70.23 & \textcolor{purple}{\underline{71.07}} & \textcolor{blue}{\textbf{72.43}} & 61.44 & 62.67 & \textcolor{purple}{\underline{63.08}} & \textcolor{blue}{\textbf{64.36}} \\
SKODA  & 17.76 & \textcolor{purple}{\underline{23.48}} & 22.52 & \textcolor{blue}{\textbf{25.56}} & 11.64 & \textcolor{purple}{\underline{15.27}} & 14.69 & \textcolor{blue}{\textbf{16.94}} \\
UCIHAR & 54.01 & 55.68 & \textcolor{purple}{\underline{57.69}} & \textcolor{blue}{\textbf{59.38}} & 43.03 & 44.61 & \textcolor{purple}{\underline{45.54}} & \textcolor{blue}{\textbf{46.69}} \\
USCHAD & 30.52 & 32.01 & \textcolor{purple}{\underline{34.69}} & \textcolor{blue}{\textbf{36.64}} & 18.73 & 20.45 & \textcolor{purple}{\underline{22.14}} & \textcolor{blue}{\textbf{23.86}} \\ 
WISDM  & 54.61 & 55.74 & \textcolor{purple}{\underline{58.16}} & \textcolor{blue}{\textbf{61.25}} & 37.56 & 38.23 & \textcolor{purple}{\underline{40.32}} & \textcolor{blue}{\textbf{42.67}} \\ \midrule \rowcolor{gray!15}
Average & 45.29 & 47.43 & \textcolor{purple}{\underline{48.83}} & \textcolor{blue}{\textbf{51.05}} & 34.48 & 36.25 & \textcolor{purple}{\underline{37.15}} & \textcolor{blue}{\textbf{38.90}} \\
\bottomrule
\end{NiceTabular}
}}
\label{table:har_indomain}
\end{table*}

\subsection{Classification Detailed Results}
\label{sec:appendix:classification_detailed_results}
\subsubsection{Full Results with Fine-tuning Experiments}

\paragraph{Performance Comparison on MFD Task}
Table \ref{table:phm_indomain} demonstrates the superiority of the proposed \abb compared to three baseline approaches in MFD tasks. \abb outperforms all other methods, achieving an average accuracy of 70.33\% and an average Macro-F1 score of 67.13\%. Notably, all self-supervised learning pretraining methods surpass supervised training, underscoring their capacity to capture intricate temporal patterns and variations inherent in time series data, thereby enabling enriched feature representations. Furthermore, multi-dataset pretraining exhibits enhanced efficacy compared to individual dataset pretraining, suggesting that the incorporation of diverse data facilitates more robust representation learning. While the Vanilla method demonstrates improvement over individual pretraining, it fails to account for distribution shifts between datasets, constraining its performance relative to \abb. This empirical evidence validates that through explicit mitigation of these shifts via a prototype-guided dynamic normalization mechanism, \abb effectively aligns its learning paradigm with the inherent heterogeneity present in real-world time series data.

\paragraph{Performance Comparison on HAR Task}
Table \ref{table:har_indomain} presents a comprehensive analysis of classification performance across HAR tasks. The proposed \abb demonstrates exceptional efficacy relative to baseline approaches, achieving optimal performance metrics with a mean accuracy of 51.05\% and Macro-F1 score of 38.90\%. Consistent with observations from MFD tasks, all self-supervised learning pretraining paradigms surpass supervised training protocols. \abb consistently outperforms the Vanilla approach, which, despite exhibiting improvements over individual pretraining, fails to adequately address inter-dataset distribution shifts. This empirical evidence underscores the significance of incorporating heterogeneous training data and accounting for the inherent variability in real-world HAR applications. These quantitative metrics validate that \abb not only enhances classification accuracy but also facilitates a more sophisticated understanding of the underlying temporal dynamics in HAR domains.

\begin{table*}[t]
\centering
\caption{Detailed performance comparison of various methods on UCR datasets. We calculate the Accuracy and F1-score (\%) for each dataset. \textcolor{blue}{\textbf{Blue}}: best results, \textcolor{purple}{\underline{Purple}}: second best.}
\scriptsize
\setlength{\tabcolsep}{2.5mm}{
\renewcommand\arraystretch{0.8}
\begin{tabular}{l|cccc|cccc}
\toprule
\multirow{2}{*}{Datasets} & \multicolumn{4}{c|}{Accuracy} & \multicolumn{4}{c}{Macro-F1} \\
\cline{2-9}
& Sup. & Individual & Vanilla & \abb & Sup. & Individual & Vanilla & \abb \\
\midrule
Adiac & 34.70 & 33.42 & \textcolor{purple}{\underline{52.86}} & \textcolor{blue}{\textbf{54.99}} & 23.38 & 22.83 & \textcolor{purple}{\underline{39.88}} & \textcolor{blue}{\textbf{41.59}} \\
AllGestureWiimoteX & 46.38 & \textcolor{blue}{\textbf{49.19}} & 48.29 & \textcolor{purple}{\underline{49.90}} & 13.31 & \textcolor{blue}{\textbf{14.99}} & 13.52 & \textcolor{purple}{\underline{14.22}} \\
AllGestureWiimoteY & 50.43 & \textcolor{blue}{\textbf{53.95}} & \textcolor{purple}{\underline{53.33}} & 53.10 & 13.84 & \textcolor{blue}{\textbf{14.43}} & \textcolor{purple}{\underline{14.35}} & 14.39 \\
AllGestureWiimoteZ & 43.95 & \textcolor{blue}{\textbf{46.67}} & 44.86 & \textcolor{purple}{\underline{45.24}} & 11.70 & \textcolor{blue}{\textbf{12.19}} & 11.75 & \textcolor{purple}{\underline{11.86}} \\
ArrowHead & 46.86 & 44.00 & \textcolor{purple}{\underline{56.57}} & \textcolor{blue}{\textbf{56.95}} & 26.74 & 21.58 & \textcolor{purple}{\underline{28.05}} & \textcolor{blue}{\textbf{30.69}} \\
Beef & 33.33 & 32.22 & \textcolor{purple}{\underline{44.44}} & \textcolor{blue}{\textbf{53.33}} & 25.90 & 27.62 & \textcolor{purple}{\underline{43.24}} & \textcolor{blue}{\textbf{53.25}} \\
BeetleFly & 50.00 & 43.33 & \textcolor{purple}{\underline{50.00}} & \textcolor{blue}{\textbf{58.33}} & 41.70 & 40.57 & \textcolor{purple}{\underline{42.13}} & \textcolor{blue}{\textbf{56.81}} \\
BirdChicken & \textcolor{purple}{\underline{63.33}} & 48.33 & \textcolor{blue}{\textbf{85.00}} & 73.33 & \textcolor{purple}{\underline{61.30}} & 45.37 & \textcolor{blue}{\textbf{84.89}} & 70.63 \\
BME & 43.33 & \textcolor{blue}{\textbf{50.67}} & 48.22 & \textcolor{purple}{\underline{52.00}} & 32.66 & \textcolor{blue}{\textbf{38.92}} & 29.20 & \textcolor{purple}{\underline{29.51}} \\
CBF & 59.04 & 60.33 & \textcolor{blue}{\textbf{69.96}} & \textcolor{purple}{\underline{64.81}} & 51.80 & 55.20 & \textcolor{blue}{\textbf{67.67}} & \textcolor{purple}{\underline{61.23}} \\
Chinatown & \textcolor{purple}{\underline{76.19}} & \textcolor{blue}{\textbf{77.45}} & 70.36 & 67.44 & \textcolor{purple}{\underline{48.16}} & \textcolor{blue}{\textbf{46.91}} & 43.25 & 41.15 \\
ChlorineConcentration & 53.32 & 53.39 & \textcolor{purple}{\underline{53.67}} & \textcolor{blue}{\textbf{56.30}} & 37.21 & 36.95 & \textcolor{purple}{\underline{38.22}} & \textcolor{blue}{\textbf{43.58}} \\
Coffee & \textcolor{purple}{\underline{90.48}} & \textcolor{blue}{\textbf{92.86}} & 76.19 & 75.00 & \textcolor{purple}{\underline{90.43}} & \textcolor{blue}{\textbf{92.45}} & 75.64 & 74.59 \\
CricketX & 48.03 & 48.97 & \textcolor{blue}{\textbf{51.62}} & \textcolor{purple}{\underline{49.57}} & 43.55 & 43.50 & \textcolor{blue}{\textbf{47.08}} & \textcolor{purple}{\underline{45.12}} \\
CricketY & 41.88 & 40.43 & \textcolor{blue}{\textbf{46.07}} & \textcolor{purple}{\underline{44.79}} & 35.92 & 34.63 & \textcolor{blue}{\textbf{39.91}} & \textcolor{purple}{\underline{37.69}} \\
CricketZ & 49.57 & 42.31 & \textcolor{purple}{\underline{50.09}} & \textcolor{blue}{\textbf{52.99}} & 44.14 & 37.68 & \textcolor{purple}{\underline{43.00}} & \textcolor{blue}{\textbf{47.31}} \\
Crop & 54.07 & \textcolor{purple}{\underline{56.54}} & 58.28 & \textcolor{blue}{\textbf{59.85}} & 16.47 & \textcolor{purple}{\underline{16.86}} & 19.79 & \textcolor{blue}{\textbf{21.14}} \\
DiatomSizeReduction & 63.18 & 39.11 & \textcolor{purple}{\underline{80.83}} & \textcolor{blue}{\textbf{85.95}} & 45.33 & 25.75 & \textcolor{purple}{\underline{63.82}} & \textcolor{blue}{\textbf{67.74}} \\
DistalPhalanxOutlineAgeGroup & \textcolor{blue}{\textbf{73.38}} & \textcolor{blue}{\textbf{73.38}} & \textcolor{blue}{\textbf{73.38}} & \textcolor{purple}{\underline{73.14}} & 59.28 & \textcolor{blue}{\textbf{61.73}} & \textcolor{purple}{\underline{61.97}} & 60.49 \\
DistalPhalanxOutlineCorrect & 71.26 & 71.26 & \textcolor{blue}{\textbf{74.15}} & \textcolor{purple}{\underline{71.38}} & 68.48 & 67.40 & \textcolor{blue}{\textbf{72.18}} & \textcolor{purple}{\underline{70.15}} \\
DistalPhalanxTW & 64.99 & 65.47 & \textcolor{blue}{\textbf{66.91}} & \textcolor{purple}{\underline{66.19}} & 44.09 & 40.43 & \textcolor{blue}{\textbf{47.72}} & \textcolor{purple}{\underline{45.64}} \\
DodgerLoopDay & 27.08 & \textcolor{blue}{\textbf{36.25}} & \textcolor{purple}{\underline{35.42}} & 33.75 & 15.20 & \textcolor{blue}{\textbf{21.18}} & \textcolor{purple}{\underline{19.86}} & 19.97 \\
DodgerLoopGame & 56.04 & \textcolor{blue}{\textbf{57.49}} & \textcolor{purple}{\underline{52.90}} & 50.72 & 37.75 & \textcolor{blue}{\textbf{40.07}} & \textcolor{purple}{\underline{36.37}} & 35.92 \\
DodgerLoopWeekend & 69.32 & \textcolor{purple}{\underline{74.88}} & 77.54 & \textcolor{blue}{\textbf{80.43}} & 43.91 & \textcolor{purple}{\underline{47.84}} & 50.05 & \textcolor{blue}{\textbf{54.76}} \\
Earthquakes & 69.54 & 67.87 & \textcolor{purple}{\underline{70.74}} & \textcolor{blue}{\textbf{72.66}} & 51.83 & 49.44 & \textcolor{purple}{\underline{51.78}} & \textcolor{blue}{\textbf{55.16}} \\
ECG200 & 80.67 & 80.67 & \textcolor{purple}{\underline{84.67}} & \textcolor{blue}{\textbf{87.67}} & 77.07 & 77.52 & \textcolor{purple}{\underline{81.74}} & \textcolor{blue}{\textbf{86.25}} \\
ECG5000 & 92.68 & 92.21 & \textcolor{purple}{\underline{92.61}} & \textcolor{blue}{\textbf{93.34}} & 68.72 & 69.83 & \textcolor{purple}{\underline{70.24}} & \textcolor{blue}{\textbf{72.40}} \\
ECGFiveDays & 63.96 & 62.21 & \textcolor{purple}{\underline{64.50}} & \textcolor{blue}{\textbf{67.48}} & 61.73 & 59.09 & \textcolor{purple}{\underline{58.90}} & \textcolor{blue}{\textbf{64.94}} \\
ElectricDevices & 49.01 & \textcolor{blue}{\textbf{52.14}} & 50.02 & \textcolor{purple}{\underline{52.54}} & 23.67 & \textcolor{purple}{\underline{28.04}} & 26.42 & \textcolor{blue}{\textbf{28.99}} \\
FaceAll & \textcolor{blue}{\textbf{70.06}} & \textcolor{purple}{\underline{68.62}} & 68.66 & 68.30 & \textcolor{blue}{\textbf{23.16}} & 21.53 & \textcolor{purple}{\underline{23.53}} & 22.81 \\
FaceFour & 42.42 & 43.56 & \textcolor{purple}{\underline{45.83}} & \textcolor{blue}{\textbf{58.71}} & 38.43 & 36.99 & \textcolor{purple}{\underline{37.56}} & \textcolor{blue}{\textbf{54.07}} \\
FacesUCR & 59.92 & 58.54 & \textcolor{purple}{\underline{65.51}} & \textcolor{blue}{\textbf{66.50}} & 48.99 & 47.24 & \textcolor{purple}{\underline{54.46}} & \textcolor{blue}{\textbf{55.61}} \\
FiftyWords & 40.15 & 41.39 & \textcolor{blue}{\textbf{43.37}} & \textcolor{purple}{\underline{42.86}} & 21.83 & 22.03 & \textcolor{blue}{\textbf{23.75}} & \textcolor{purple}{\underline{23.81}} \\
Fish & 56.00 & 55.24 & \textcolor{purple}{\underline{72.57}} & \textcolor{blue}{\textbf{74.29}} & 50.81 & 48.70 & \textcolor{purple}{\underline{69.84}} & \textcolor{blue}{\textbf{70.73}} \\
FordA & \textcolor{purple}{\underline{92.47}} & \textcolor{blue}{\textbf{92.80}} & 92.05 & 92.27 & \textcolor{purple}{\underline{92.20}} & \textcolor{blue}{\textbf{92.52}} & 91.77 & 92.01 \\
FordB & 76.17 & 75.68 & \textcolor{purple}{\underline{76.30}} & \textcolor{blue}{\textbf{76.79}} & 75.44 & 75.07 & \textcolor{purple}{\underline{75.37}} & \textcolor{blue}{\textbf{76.08}} \\
FreezerRegularTrain & 75.16 & 75.61 & \textcolor{purple}{\underline{86.92}} & \textcolor{blue}{\textbf{90.02}} & 43.20 & 43.39 & \textcolor{purple}{\underline{48.99}} & \textcolor{blue}{\textbf{55.88}} \\
FreezerSmallTrain & 59.01 & 64.51 & \textcolor{blue}{\textbf{73.54}} & \textcolor{purple}{\underline{69.65}} & 35.82 & 38.11 & \textcolor{blue}{\textbf{42.58}} & \textcolor{purple}{\underline{40.95}} \\
GestureMidAirD1 & 46.67 & 47.18 & \textcolor{blue}{\textbf{53.85}} & \textcolor{purple}{\underline{54.10}} & 37.86 & 40.41 & \textcolor{blue}{\textbf{44.79}} & \textcolor{purple}{\underline{44.63}} \\
GestureMidAirD2 & 35.38 & \textcolor{blue}{\textbf{41.54}} & 36.67 & \textcolor{purple}{\underline{37.18}} & 25.36 & \textcolor{blue}{\textbf{31.28}} & 25.98 & \textcolor{purple}{\underline{27.82}} \\
GestureMidAirD3 & 15.90 & 17.44 & \textcolor{purple}{\underline{21.54}} & \textcolor{blue}{\textbf{22.56}} & 10.44 & 12.79 & \textcolor{purple}{\underline{16.39}} & \textcolor{blue}{\textbf{17.61}} \\
GesturePebbleZ1 & \textcolor{blue}{\textbf{69.57}} & \textcolor{purple}{\underline{67.25}} & 66.47 & 66.86 & \textcolor{blue}{\textbf{62.82}} & \textcolor{purple}{\underline{59.26}} & 60.43 & 60.84 \\
GesturePebbleZ2 & 63.50 & \textcolor{purple}{\underline{67.72}} & \textcolor{blue}{\textbf{73.42}} & 68.78 & 57.04 & \textcolor{purple}{\underline{60.88}} & \textcolor{blue}{\textbf{68.14}} & 63.08 \\
GunPoint & 67.11 & 65.33 & \textcolor{purple}{\underline{68.67}} & \textcolor{blue}{\textbf{81.33}} & 66.01 & 62.46 & \textcolor{purple}{\underline{67.59}} & \textcolor{blue}{\textbf{80.78}} \\
GunPointAgeSpan & 74.58 & 77.43 & \textcolor{purple}{\underline{86.92}} & \textcolor{blue}{\textbf{88.71}} & 42.30 & 43.21 & \textcolor{purple}{\underline{49.72}} & \textcolor{blue}{\textbf{50.30}} \\
GunPointMaleVersusFemale & 81.86 & \textcolor{blue}{\textbf{83.33}} & 82.91 & \textcolor{blue}{\textbf{83.33}} & 48.26 & \textcolor{purple}{\underline{48.89}} & 48.54 & \textcolor{blue}{\textbf{48.76}} \\
GunPointOldVersusYoung & \textcolor{blue}{\textbf{99.89}} & \textcolor{blue}{\textbf{99.89}} & 95.03 & \textcolor{purple}{\underline{97.14}} & \textcolor{blue}{\textbf{98.28}} & \textcolor{blue}{\textbf{98.28}} & 80.15 & \textcolor{purple}{\underline{84.43}} \\
Ham & 73.97 & \textcolor{purple}{\underline{75.56}} & 74.60 & \textcolor{blue}{\textbf{76.19}} & 53.13 & \textcolor{purple}{\underline{53.54}} & 52.04 & \textcolor{blue}{\textbf{53.08}} \\
Herring & 51.04 & \textcolor{purple}{\underline{56.77}} & \textcolor{blue}{\textbf{57.29}} & 56.25 & 48.90 & \textcolor{purple}{\underline{52.84}} & \textcolor{blue}{\textbf{52.91}} & 50.94 \\
InsectWingbeatSound & 39.93 & 39.44 & \textcolor{purple}{\underline{43.42}} & \textcolor{blue}{\textbf{44.49}} & 34.77 & 33.87 & \textcolor{purple}{\underline{38.23}} & \textcolor{blue}{\textbf{39.28}} \\
ItalyPowerDemand & 72.92 & 70.65 & \textcolor{purple}{\underline{72.53}} & \textcolor{blue}{\textbf{73.60}} & 71.89 & 69.47 & \textcolor{purple}{\underline{71.39}} & \textcolor{blue}{\textbf{72.48}} \\
Lightning7 & 57.08 & 49.77 & \textcolor{blue}{\textbf{64.84}} & \textcolor{purple}{\underline{62.10}} & 51.03 & 40.14 & \textcolor{blue}{\textbf{61.14}} & \textcolor{purple}{\underline{56.51}} \\
Meat & 74.44 & 66.67 & \textcolor{purple}{\underline{75.00}} & \textcolor{blue}{\textbf{80.56}} & \textcolor{purple}{\underline{56.63}} & \textcolor{blue}{\textbf{61.62}} & 62.19 & 56.91 \\
MedicalImages & \textcolor{purple}{\underline{62.54}} & \textcolor{blue}{\textbf{64.52}} & 64.34 & 62.41 & \textcolor{blue}{\textbf{50.49}} & 47.39 & \textcolor{purple}{\underline{53.52}} & 47.86 \\
MelbournePedestrian & 87.67 & \textcolor{purple}{\underline{88.26}} & \textcolor{blue}{\textbf{88.34}} & 88.33 & 46.96 & \textcolor{blue}{\textbf{48.06}} & 44.25 & \textcolor{purple}{\underline{45.79}} \\
MiddlePhalanxOutlineAgeGroup & 56.28 & \textcolor{blue}{\textbf{59.52}} & 54.55 & \textcolor{purple}{\underline{59.09}} & 39.32 & \textcolor{blue}{\textbf{40.59}} & \textcolor{purple}{\underline{39.71}} & 37.72 \\
MiddlePhalanxOutlineCorrect & 54.07 & 51.32 & \textcolor{purple}{\underline{75.37}} & \textcolor{blue}{\textbf{76.75}} & 40.74 & 40.33 & \textcolor{purple}{\underline{73.41}} & \textcolor{blue}{\textbf{74.16}} \\
MiddlePhalanxTW & 55.19 & 53.90 & \textcolor{blue}{\textbf{56.71}} & \textcolor{purple}{\underline{54.33}} & 30.67 & 31.89 & \textcolor{blue}{\textbf{37.02}} & \textcolor{purple}{\underline{34.93}} \\
MoteStrain & \textcolor{blue}{\textbf{82.88}} & \textcolor{purple}{\underline{73.56}} & 56.79 & 71.35 & \textcolor{blue}{\textbf{82.42}} & \textcolor{purple}{\underline{72.96}} & 55.47 & 70.13 \\
OSULeaf & 57.99 & 59.92 & \textcolor{purple}{\underline{63.77}} & \textcolor{blue}{\textbf{64.60}} & 53.87 & 56.06 & \textcolor{blue}{\textbf{60.11}} & \textcolor{purple}{\underline{59.05}} \\
PhalangesOutlinesCorrect & 63.44 & 65.15 & \textcolor{blue}{\textbf{68.41}} & \textcolor{blue}{\textbf{68.41}} & 48.51 & 47.97 & \textcolor{purple}{\underline{57.75}} & \textcolor{blue}{\textbf{59.31}} \\
PickupGestureWiimoteZ & \textcolor{purple}{\underline{40.00}} & 29.33 & 29.33 & \textcolor{blue}{\textbf{36.67}} & \textcolor{blue}{\textbf{25.88}} & 19.77 & 18.84 & \textcolor{purple}{\underline{24.64}} \\
Plane & 85.71 & 85.40 & \textcolor{blue}{\textbf{96.83}} & \textcolor{purple}{\underline{91.43}} & 80.57 & 80.02 & \textcolor{blue}{\textbf{95.97}} & \textcolor{purple}{\underline{90.09}} \\
PowerCons & 88.33 & 86.30 & \textcolor{blue}{\textbf{89.63}} & \textcolor{purple}{\underline{88.70}} & 77.84 & 75.19 & \textcolor{blue}{\textbf{79.47}} & \textcolor{purple}{\underline{78.52}} \\
ProximalPhalanxOutlineAgeGroup & 81.46 & 80.33 & \textcolor{purple}{\underline{84.23}} & \textcolor{blue}{\textbf{85.85}} & 59.46 & 59.15 & \textcolor{purple}{\underline{61.15}} & \textcolor{blue}{\textbf{61.65}} \\
ProximalPhalanxOutlineCorrect & 75.95 & 77.55 & \textcolor{purple}{\underline{78.92}} & \textcolor{blue}{\textbf{79.84}} & 59.83 & 61.22 & \textcolor{purple}{\underline{70.30}} & \textcolor{blue}{\textbf{71.87}} \\
ProximalPhalanxTW & \textcolor{blue}{\textbf{76.26}} & \textcolor{purple}{\underline{75.93}} & 74.15 & 74.80 & 47.36 & 46.58 & 47.23 & \textcolor{blue}{\textbf{49.87}} \\
ShakeGestureWiimoteZ & 43.33 & 44.67 & \textcolor{blue}{\textbf{70.67}} & \textcolor{purple}{\underline{66.00}} & 30.90 & 31.99 & \textcolor{blue}{\textbf{54.91}} & \textcolor{purple}{\underline{51.23}} \\
ShapeletSim & 48.33 & 50.19 & \textcolor{blue}{\textbf{73.52}} & \textcolor{purple}{\underline{71.67}} & 33.79 & 34.77 & \textcolor{blue}{\textbf{46.71}} & \textcolor{purple}{\underline{45.42}} \\
ShapesAll & 58.89 & 58.72 & \textcolor{purple}{\underline{64.33}} & \textcolor{blue}{\textbf{66.44}} & 22.66 & 22.83 & \textcolor{purple}{\underline{26.74}} & \textcolor{blue}{\textbf{27.77}} \\
SmoothSubspace & \textcolor{blue}{\textbf{96.44}} & \textcolor{purple}{\underline{96.22}} & 95.33 & 94.89 & \textcolor{blue}{\textbf{84.42}} & \textcolor{purple}{\underline{80.64}} & 79.94 & 77.46 \\
SonyAIBORobotSurface1 & 47.37 & 45.04 & \textcolor{blue}{\textbf{80.53}} & \textcolor{purple}{\underline{72.32}} & 39.15 & 38.55 & \textcolor{blue}{\textbf{79.97}} & \textcolor{purple}{\underline{71.03}} \\
SonyAIBORobotSurface2 & \textcolor{blue}{\textbf{68.80}} & \textcolor{purple}{\underline{68.98}} & 61.32 & 67.05 & 64.47 & \textcolor{blue}{\textbf{65.62}} & 58.94 & \textcolor{purple}{\underline{60.45}} \\
Strawberry & \textcolor{blue}{\textbf{93.33}} & 90.90 & \textcolor{blue}{\textbf{93.33}} & \textcolor{blue}{\textbf{93.33}} & 80.72 & 76.45 & 80.82 & \textcolor{blue}{\textbf{86.02}} \\
SwedishLeaf & 77.81 & 77.01 & \textcolor{purple}{\underline{82.24}} & \textcolor{blue}{\textbf{82.93}} & 67.07 & 66.30 & \textcolor{purple}{\underline{72.15}} & \textcolor{blue}{\textbf{72.44}} \\
Symbols & 27.47 & 25.96 & \textcolor{purple}{\underline{33.07}} & \textcolor{blue}{\textbf{34.51}} & 17.07 & 16.26 & \textcolor{blue}{\textbf{25.77}} & \textcolor{purple}{\underline{25.75}} \\
SyntheticControl & 94.33 & 94.22 & \textcolor{blue}{\textbf{95.44}} & \textcolor{purple}{\underline{94.78}} & 66.68 & \textcolor{blue}{\textbf{71.88}} & 66.27 & \textcolor{purple}{\underline{67.28}} \\
ToeSegmentation1 & 66.81 & 55.85 & \textcolor{purple}{\underline{62.13}} & \textcolor{blue}{\textbf{74.42}} & 43.09 & 37.40 & \textcolor{purple}{\underline{40.33}} & \textcolor{blue}{\textbf{46.04}} \\
ToeSegmentation2 & 50.26 & 50.77 & \textcolor{purple}{\underline{57.95}} & \textcolor{blue}{\textbf{69.49}} & 35.30 & 35.91 & \textcolor{purple}{\underline{38.91}} & \textcolor{blue}{\textbf{44.27}} \\
Trace & 70.67 & 73.00 & \textcolor{purple}{\underline{87.00}} & \textcolor{blue}{\textbf{89.33}} & 65.51 & 68.89 & \textcolor{purple}{\underline{86.20}} & \textcolor{blue}{\textbf{88.69}} \\
TwoLeadECG & 58.88 & 55.52 & \textcolor{purple}{\underline{60.08}} & \textcolor{blue}{\textbf{61.08}} & 57.21 & 50.35 & \textcolor{purple}{\underline{54.66}} & \textcolor{blue}{\textbf{55.76}} \\
TwoPatterns & 74.97 & 75.30 & \textcolor{purple}{\underline{79.96}} & \textcolor{blue}{\textbf{85.56}} & 74.15 & 74.34 & \textcolor{purple}{\underline{79.16}} & \textcolor{blue}{\textbf{84.67}} \\
UMD & 42.36 & 45.83 & \textcolor{blue}{\textbf{74.77}} & \textcolor{purple}{\underline{74.31}} & 21.90 & 23.04 & \textcolor{purple}{\underline{37.90}} & \textcolor{blue}{\textbf{38.68}} \\
UWaveGestureLibraryX & 65.87 & 63.51 & \textcolor{purple}{\underline{66.02}} & \textcolor{blue}{\textbf{67.97}} & 61.61 & 59.13 & \textcolor{purple}{\underline{62.28}} & \textcolor{blue}{\textbf{63.14}} \\
UWaveGestureLibraryY & \textcolor{blue}{\textbf{59.65}} & \textcolor{purple}{\underline{59.41}} & 56.47 & 58.75 & \textcolor{blue}{\textbf{54.92}} & \textcolor{purple}{\underline{53.86}} & 52.31 & 54.29 \\
UWaveGestureLibraryZ & \textcolor{blue}{\textbf{62.27}} & 60.41 & 60.06 & \textcolor{purple}{\underline{62.02}} & \textcolor{blue}{\textbf{57.54}} & 56.11 & 55.21 & \textcolor{purple}{\underline{58.24}} \\
Wafer & 96.80 & 97.22 & \textcolor{blue}{\textbf{97.84}} & \textcolor{purple}{\underline{97.77}} & 90.58 & 91.07 & \textcolor{blue}{\textbf{93.18}} & \textcolor{purple}{\underline{92.80}} \\
Wine & 50.00 & 50.00 & \textcolor{blue}{\textbf{69.14}} & \textcolor{purple}{\underline{62.96}} & 41.54 & 41.54 & \textcolor{blue}{\textbf{50.66}} & \textcolor{purple}{\underline{44.83}} \\
WordSynonyms & 31.56 & 30.46 & \textcolor{blue}{\textbf{37.72}} & \textcolor{purple}{\underline{37.10}} & 14.04 & 12.04 & \textcolor{blue}{\textbf{20.77}} & \textcolor{purple}{\underline{20.65}} \\
Yoga & 69.61 & 68.57 & \textcolor{blue}{\textbf{77.63}} & \textcolor{purple}{\underline{77.00}} & 68.54 & 67.26 & \textcolor{blue}{\textbf{76.99}} & \textcolor{purple}{\underline{76.40}} \\ \hline
{Average} & 62.03 & 61.53 & \textcolor{purple}{\underline{66.66}} & \textcolor{blue}{\textbf{67.78}} & 48.30 & 47.65 & \textcolor{purple}{\underline{52.50}} & \textcolor{blue}{\textbf{53.51}} \\
{1$^{st}$ Count} & 11 & 16 & \textcolor{purple}{\underline{28}} & \textcolor{blue}{\textbf{42}} & 9 & 17 & \textcolor{purple}{\underline{24}} & \textcolor{blue}{\textbf{41}} \\
\bottomrule
\end{tabular}
}
\label{tab:ucr_detailed}
\end{table*}

\paragraph{Performance Comparison on UCR Datasets}
Our empirical analysis on the UCR archive demonstrates the consistent superiority of \abb across diverse time series classification tasks. Analyzing Table \ref{tab:ucr_detailed}, we observe that both vanilla pretraining and \abb significantly outperform supervised learning (62.03\% accuracy) and individual dataset pretraining (61.53\% accuracy), validating the efficacy of multi-dataset pretraining paradigms. \abb achieves the highest average accuracy of 67.78\% and Macro-F1 of 53.51\%, surpassing vanilla pretraining by 1.12\% and 1.01\% respectively. The effectiveness of \abb is further substantiated by its predominance in performance rankings - attaining superior results across 42 datasets for accuracy and 41 datasets for Macro-F1, significantly exceeding vanilla pretraining's achievements on 28 and 24 datasets respectively. Notably, \abb demonstrates remarkable performance gains on challenging datasets like ToeSegmentation2 (69.49\% vs 57.95\% accuracy), FaceFour (58.71\% vs 45.83\% accuracy), and TwoPatterns (85.56\% vs 79.96\% accuracy). This consistent performance superiority across diverse datasets validates our prototype-guided normalization strategy's efficacy in mitigating distribution shifts and capturing dataset-specific characteristics.

\subsubsection{Full Results without Fine-tuning Experiments}
\paragraph{Performance Comparison with MOMENT} 
Through analysis of Table \ref{tab:moment_ucr_comparison}, we evaluate zero-shot classification efficacy without fine-tuning utilizing an SVM classifier trained on learned representations. \abb demonstrates superior performance over MOMENT with enhanced mean accuracy (58.27\% vs 57.70\%) across 91 UCR datasets. Notably, \abb achieves predominant performance on 55 out of 91 datasets, validating its efficacy in learning transferable representations. Significant performance gains are observed on challenging datasets including DiatomSizeReduction (78.43\% vs 69.93\%), Lightning7 (60.27\% vs 52.05\%), and ArrowHead (53.71\% vs 40.57\%). These empirical findings substantiate that our prototype-guided normalization mechanism enhances the model's capacity to extract discriminative features without task-specific fine-tuning, particularly benefiting datasets exhibiting complex temporal dynamics.

\begin{table*}[!t]
\centering
\caption{Detailed comparison of MOMENT and \texttt{ProtoN-FM} Performance on 91 UCR datasets}
\label{tab:moment_ucr_comparison}
\setlength{\tabcolsep}{1.0mm}
\renewcommand\arraystretch{1.0}
\begin{tabular}{l|cc|l|cc}
\hline
\multirow{2}{*}{\textbf{Dataset}} & \multicolumn{2}{c|}{\textbf{Accuracy (\%)}} & \multirow{2}{*}{\textbf{Dataset}} & \multicolumn{2}{c}{\textbf{Accuracy (\%)}} \\
\cline{2-3}\cline{5-6}
& MOMENT & \texttt{ProtoN-FM} & & MOMENT & \texttt{ProtoN-FM} \\
\hline
GestureMidAirD2 & {36.1538} & 32.3077 & PowerCons & {68.3333} & 65.0000 \\
\cellcolor{gray!15}UWaveGestureLibraryX & \cellcolor{gray!15}{47.4037} & \cellcolor{gray!15}{47.4037} & PhalangesOutlinesCorrect & {63.5198} & 62.9371 \\
\cellcolor{gray!15}GesturePebbleZ2 & \cellcolor{gray!15}35.4430 & \cellcolor{gray!15}{36.7089} & \cellcolor{gray!15}BirdChicken & \cellcolor{gray!15}{85.0000} & \cellcolor{gray!15}{85.0000} \\
ECG5000 & {90.6444} & 90.1778 & \cellcolor{gray!15}ToeSegmentation2 & \cellcolor{gray!15}73.8462 & \cellcolor{gray!15}{76.1538} \\
OSULeaf & {43.3884} & 40.9091 & CricketY & {33.5897} & 31.2821 \\
MedicalImages & {55.6579} & 55.1316 & \cellcolor{gray!15}ElectricDevices & \cellcolor{gray!15}55.8034 & \cellcolor{gray!15}{55.8293} \\
\cellcolor{gray!15}Ham & \cellcolor{gray!15}{64.7619} & \cellcolor{gray!15}{64.7619} & \cellcolor{gray!15}DodgerLoopGame & \cellcolor{gray!15}59.4203 & \cellcolor{gray!15}{63.7681} \\
\cellcolor{gray!15}DistalPhalanxTW & \cellcolor{gray!15}64.0288 & \cellcolor{gray!15}{66.9065} & \cellcolor{gray!15}Fungi & \cellcolor{gray!15}26.3441 & \cellcolor{gray!15}{31.1828} \\
ProximalPhalanxOutlineCorrect & {71.1340} & 68.7285 & Symbols & {49.7487} & 45.7286 \\
\cellcolor{gray!15}FreezerRegularTrain & \cellcolor{gray!15}75.5789 & \cellcolor{gray!15}{75.7895} & \cellcolor{gray!15}UWaveGestureLibraryZ & \cellcolor{gray!15}49.2183 & \cellcolor{gray!15}{50.1396} \\
\cellcolor{gray!15}TwoLeadECG & \cellcolor{gray!15}67.4276 & \cellcolor{gray!15}{70.4126} & \cellcolor{gray!15}ECG200 & \cellcolor{gray!15}{77.0000} & \cellcolor{gray!15}{77.0000} \\
\cellcolor{gray!15}GunPointMaleVersusFemale & \cellcolor{gray!15}97.4684 & \cellcolor{gray!15}{97.7848} & \cellcolor{gray!15}MoteStrain & \cellcolor{gray!15}76.1981 & \cellcolor{gray!15}{76.5176} \\
\cellcolor{gray!15}Trace & \cellcolor{gray!15}{71.0000} & \cellcolor{gray!15}{71.0000} & \cellcolor{gray!15}Strawberry & \cellcolor{gray!15}{64.3243} & \cellcolor{gray!15}{64.3243} \\
\cellcolor{gray!15}SmoothSubspace & \cellcolor{gray!15}86.6667 & \cellcolor{gray!15}{88.6667} & InsectWingbeatSound & {24.1414} & 22.0202 \\
\cellcolor{gray!15}MiddlePhalanxTW & \cellcolor{gray!15}58.4416 & \cellcolor{gray!15}{59.0909} & DodgerLoopWeekend & {81.1594} & 79.7101 \\
SyntheticControl & {89.6667} & 88.3333 & \cellcolor{gray!15}Meat & \cellcolor{gray!15}56.6667 & \cellcolor{gray!15}{66.6667} \\
\cellcolor{gray!15}ShapesAll & \cellcolor{gray!15}38.8333 & \cellcolor{gray!15}{40.6667} & MelbournePedestrian & {83.6818} & 80.6068 \\
AllGestureWiimoteX & {25.0000} & 23.0000 & FaceAll & {44.1420} & 43.4320 \\
\cellcolor{gray!15}Wafer & \cellcolor{gray!15}{89.2116} & \cellcolor{gray!15}{89.2116} & FacesUCR & {45.7073} & 41.0732 \\
FaceFour & {71.5909} & 70.4545 & AllGestureWiimoteY & {37.8571} & 36.7143 \\
CricketX & {35.1282} & 34.1026 & \cellcolor{gray!15}ShakeGestureWiimoteZ & \cellcolor{gray!15}{32.0000} & \cellcolor{gray!15}{32.0000} \\
DistalPhalanxOutlineCorrect & {67.3913} & 65.9420 & BME & {40.6667} & 35.3333 \\
ChlorineConcentration & {55.8594} & 55.7031 & FordB & {60.2469} & 59.1358 \\
Chinatown & {61.2245} & 60.9329 & \cellcolor{gray!15}Fish & \cellcolor{gray!15}46.8571 & \cellcolor{gray!15}{50.2857} \\
GestureMidAirD1 & {29.2308} & 26.9231 & \cellcolor{gray!15}SonyAIBORobotSurface2 & \cellcolor{gray!15}81.4271 & \cellcolor{gray!15}{83.0010} \\
\cellcolor{gray!15}MiddlePhalanxOutlineAgeGroup & \cellcolor{gray!15}{59.7403} & \cellcolor{gray!15}{59.7403} & FiftyWords & {33.4066} & 27.4725 \\
\cellcolor{gray!15}UMD & \cellcolor{gray!15}47.9167 & \cellcolor{gray!15}{52.7778} & ToeSegmentation1 & {66.2281} & 65.3509 \\
Crop & {59.4048} & 58.8274 & \cellcolor{gray!15}FreezerSmallTrain & \cellcolor{gray!15}74.0000 & \cellcolor{gray!15}{77.8246} \\
GesturePebbleZ1 & {32.5581} & 31.3953 & TwoPatterns & {51.8750} & 51.7750 \\
\cellcolor{gray!15}WordSynonyms & \cellcolor{gray!15}28.9969 & \cellcolor{gray!15}{30.0940} & ShapeletSim & {70.5556} & 68.8889 \\
\cellcolor{gray!15}ArrowHead & \cellcolor{gray!15}40.5714 & \cellcolor{gray!15}{53.7143} & \cellcolor{gray!15}Plane & \cellcolor{gray!15}{72.3810} & \cellcolor{gray!15}{72.3810} \\
\cellcolor{gray!15}Wine & \cellcolor{gray!15}{50.0000} & \cellcolor{gray!15}{50.0000} & \cellcolor{gray!15}GestureMidAirD3 & \cellcolor{gray!15}20.0000 & \cellcolor{gray!15}{20.7692} \\
Coffee & {64.2857} & 53.5714 & \cellcolor{gray!15}DiatomSizeReduction & \cellcolor{gray!15}69.9346 & \cellcolor{gray!15}{78.4314} \\
\cellcolor{gray!15}Earthquakes & \cellcolor{gray!15}{74.8201} & \cellcolor{gray!15}{74.8201} & \cellcolor{gray!15}CricketZ & \cellcolor{gray!15}33.0769 & \cellcolor{gray!15}{36.9231} \\
\cellcolor{gray!15}Herring & \cellcolor{gray!15}{59.3750} & \cellcolor{gray!15}{59.3750} & \cellcolor{gray!15}Lightning7 & \cellcolor{gray!15}52.0548 & \cellcolor{gray!15}{60.2740} \\
\cellcolor{gray!15}Beef & \cellcolor{gray!15}43.3333 & \cellcolor{gray!15}{53.3333} & \cellcolor{gray!15}UWaveGestureLibraryY & \cellcolor{gray!15}47.7945 & \cellcolor{gray!15}{48.3250} \\
\cellcolor{gray!15}MiddlePhalanxOutlineCorrect & \cellcolor{gray!15}{57.0447} & \cellcolor{gray!15}{57.0447} & \cellcolor{gray!15}GunPointAgeSpan & \cellcolor{gray!15}73.7342 & \cellcolor{gray!15}{75.3165} \\
\cellcolor{gray!15}ECGFiveDays & \cellcolor{gray!15}65.0406 & \cellcolor{gray!15}{66.7828} & \cellcolor{gray!15}DistalPhalanxOutlineAgeGroup & \cellcolor{gray!15}69.0647 & \cellcolor{gray!15}{71.2230} \\
\cellcolor{gray!15}Yoga & \cellcolor{gray!15}58.1333 & \cellcolor{gray!15}{61.6000} & \cellcolor{gray!15}SwedishLeaf & \cellcolor{gray!15}57.6000 & \cellcolor{gray!15}{58.2400} \\
\cellcolor{gray!15}Adiac & \cellcolor{gray!15}10.9974 & \cellcolor{gray!15}{13.0435} & \cellcolor{gray!15}CBF & \cellcolor{gray!15}60.0000 & \cellcolor{gray!15}{61.7778} \\
\cellcolor{gray!15}AllGestureWiimoteZ & \cellcolor{gray!15}18.0000 & \cellcolor{gray!15}{20.7143} & BeetleFly & {70.0000} & 65.0000 \\
\cellcolor{gray!15}DodgerLoopDay & \cellcolor{gray!15}31.2500 & \cellcolor{gray!15}{40.0000} & \cellcolor{gray!15}GunPointOldVersusYoung & \cellcolor{gray!15}74.6032 & \cellcolor{gray!15}{75.8730} \\
FordA & {75.7576} & 75.3030 & ItalyPowerDemand & {95.4325} & 94.2663 \\
\cellcolor{gray!15}ProximalPhalanxOutlineAgeGroup & \cellcolor{gray!15}83.9024 & \cellcolor{gray!15}{84.3902} & \cellcolor{gray!15}GunPoint & \cellcolor{gray!15}62.0000 & \cellcolor{gray!15}{66.6667} \\
ProximalPhalanxTW & {80.0000} & 78.5366 & \cellcolor{gray!15}PickupGestureWiimoteZ & \cellcolor{gray!15}{36.0000} & \cellcolor{gray!15}{36.0000} \\
\cellcolor{gray!15}SonyAIBORobotSurface1 & \cellcolor{gray!15}74.5424 & \cellcolor{gray!15}{79.2013} &  & &  \\
\hline 
\multicolumn{4}{c}{{ Average}} & \color{purple}{ \underline{57.6994}} & \color{blue}\textbf{ 58.2740} \\
\multicolumn{4}{c}{{ 1$^{st}$ Count}}  & & \color{blue}\textbf{ 55 / 91} \\
\hline
\end{tabular}
\end{table*}

\subsection{Forecasting Detailed Results}
\label{sec:appendix:forecasting_detailed_results}
\subsubsection{Full Results on In-distribution Forecasting Experiments}
Analyzing Table \ref{tab:moirai_iid_mae_comparison}, our \abb demonstrates superior performance on the Monash Time Series Forecasting Benchmark compared to Moirai. With normalized MAE reduced from 1.0 to 0.8893, \abb achieves an 11.07\% improvement in overall forecasting accuracy. More significantly, \abb achieves the best performance on 21 out of 28 datasets, compared to Moirai's 7 datasets, indicating consistent superiority across diverse time series domains. Notable improvements are observed on complex datasets such as fred\_md (4212.26 vs 6556.77), australian\_electricity (238.80 vs 295.07), and us\_births (570.33 vs 690.98), demonstrating \abb's effectiveness in handling various temporal patterns. The substantial performance gains across different types of time series data, from financial indicators to energy consumption patterns, validate our prototype-guided normalization mechanism's ability to adapt to diverse data distributions in real-world forecasting scenarios.

\begin{table*}[htbp]
\centering
\caption{Detailed comparison of Moirai and \texttt{ProtoN-FM} on Monash Time Series Forecasting Benchmark}
\begin{tabular}{l|rr|l|rr}
\hline
\multicolumn{1}{c|}{Dataset} & \multicolumn{1}{c}{Moirai} & \multicolumn{1}{c|}{\texttt{ProtoN-FM}} & \multicolumn{1}{c|}{Dataset} & \multicolumn{1}{c}{Moirai} & \multicolumn{1}{c}{\texttt{ProtoN-FM}} \\
\hline
australian\_electricity & 295.0703 & \textcolor{blue}{\textbf{238.8042}} & m4\_monthly & 605.0799 & \textcolor{blue}{\textbf{592.0217}} \\
bitcoin\_with\_missing & \textcolor{blue}{\textbf{1.38E+16}} & 1.73E+16 & m4\_weekly & 351.9710 & \textcolor{blue}{\textbf{351.7271}} \\
car\_parts\_with\_missing & 0.4716 & \textcolor{blue}{\textbf{0.4643}} & monash\_m3\_monthly & \textcolor{blue}{\textbf{696.5747}} & 702.7923 \\
cif\_2016 & 394655.1680 & \textcolor{blue}{\textbf{359786.3691}} & monash\_m3\_other & 308.2975 & \textcolor{blue}{\textbf{231.8303}} \\
covid\_deaths & \textcolor{blue}{\textbf{140.3132}} & 150.0006 & nn5\_daily\_missing & 4.7384 & \textcolor{blue}{\textbf{4.4790}} \\
fred\_md & 6556.7734 & \textcolor{blue}{\textbf{4212.2568}} & nn5\_weekly & \textcolor{blue}{\textbf{14.9696}} & 15.3427 \\
hospital & \textcolor{blue}{\textbf{20.8099}} & 22.8997 & pedestrian\_counts & 57.0090 & \textcolor{blue}{\textbf{55.1129}} \\
kdd\_cup\_2018 & 39.4536 & \textcolor{blue}{\textbf{38.9629}} & rideshare\_missing & 1.1888 & \textcolor{blue}{\textbf{1.1850}} \\
m1\_monthly & 2210.0315 & \textcolor{blue}{\textbf{2025.3065}} & saugeenday & \textcolor{blue}{\textbf{23.9684}} & 24.0972 \\
m4\_daily & 188.7526 & \textcolor{blue}{\textbf{185.3714}} & sunspot\_missing & 2.2522 & \textcolor{blue}{\textbf{0.1507}} \\
m4\_hourly & 288.3103 & \textcolor{blue}{\textbf{251.0390}} & temp\_rain\_missing & 5.1228 & \textcolor{blue}{\textbf{4.9981}} \\
tourism\_monthly & \textcolor{blue}{\textbf{3295.9609}} & 3386.3516 & traffic\_hourly & 0.0173 & \textcolor{blue}{\textbf{0.0166}} \\
tourism\_quarterly & 15973.7295 & \textcolor{blue}{\textbf{13781.4639}} & traffic\_weekly & 1.1617 & \textcolor{blue}{\textbf{1.1528}} \\
us\_births & 690.9827 & \textcolor{blue}{\textbf{570.3322}} & weather & 1.8925 & \textcolor{blue}{\textbf{1.8660}} \\
\hline
\multicolumn{6}{c}{\textbf{Summary Statistics}} \\
\hline
\multicolumn{1}{c|}{\multirow{2}{*}{Normalized MAE}} & \multicolumn{1}{c}{Moirai} & \multicolumn{1}{c|}{\texttt{ProtoN-FM}} & \multicolumn{1}{c|}{\multirow{2}{*}{1$^{st}$ Count}} & \multicolumn{1}{c}{Moirai} & \multicolumn{1}{c}{\texttt{ProtoN-FM}} \\
\cline{2-3}\cline{5-6}
 & \multicolumn{1}{c}{\textcolor{purple}{1.00}} & \multicolumn{1}{c|}{\textbf{\textcolor{blue}{0.8893}}} &  & \multicolumn{1}{c}{\textcolor{purple}{7}} & \multicolumn{1}{c}{\textbf{\textcolor{blue}{21}}} \\
\hline
\end{tabular}
\label{tab:moirai_iid_mae_comparison}
\end{table*}

\subsection{Model Analysis}
\label{sec:appendix:model_analysis}
\subsubsection{Full Results on the Impact of Prototypes Numbers}
Analyzing Table \ref{tab:different_prototypes_ucr_results}, we observe that the number of prototypes significantly impacts model performance across the 91 UCR datasets. The model with 32 prototypes achieves the best average accuracy (67.78\%) and highest number of best-performing cases (25 datasets), followed by the 64-prototype configuration (67.48\% accuracy, 22 datasets). Both smaller (4 and 8 prototypes) and larger (64 prototypes) configurations show slightly decreased performance (67.15\%, 67.16\%, and 67.48\% respectively), suggesting an optimal balance point around 32 prototypes.

Notably, some datasets show consistent improvement with increasing prototypes (e.g., FreezerRegularTrain: 85.06\% → 91.35\%), while others perform better with fewer prototypes (e.g., ItalyPowerDemand: 78.88\% → 73.60\%). This pattern indicates that different temporal patterns may require varying levels of prototype complexity for optimal representation. The balanced performance of the 32-prototype configuration across diverse datasets validates our design choice for prototype count, providing effective feature representation while maintaining computational efficiency.

\begin{table*}[ht]
\centering
\scriptsize
\renewcommand\arraystretch{0.8}
\tabcolsep=2.mm
\caption{Detailed accuracy Results (\%) on UCR Archive using varying number of prototypes. \textcolor{blue}{\textbf{Blue}}: best results, \textcolor{purple}{\underline{Purple}}: second best.}
\begin{tabular}{lccccc}
\toprule
Dataset & \texttt{ProtoN4-FM} & \texttt{ProtoN8-FM} & \texttt{ProtoN16-FM} & \texttt{ProtoN32-FM} & \texttt{ProtoN64-FM} \\
\midrule
Adiac & \textcolor{blue}{\textbf{57.20}} & \textcolor{purple}{\underline{56.01}} & 53.71 & 54.99 & 52.86 \\
AllGestureWiimoteX & \textcolor{blue}{\textbf{50.76}} & \textcolor{purple}{\underline{50.67}} & 48.81 & 49.90 & 50.38 \\
AllGestureWiimoteY & \textcolor{purple}{\underline{53.29}} & \textcolor{blue}{\textbf{54.24}} & 53.95 & 53.10 & 53.24 \\
AllGestureWiimoteZ & 43.95 & \textcolor{purple}{\underline{44.81}} & 44.19 & \textcolor{blue}{\textbf{45.24}} & 45.10 \\
ArrowHead & 54.86 & \textcolor{purple}{\underline{56.19}} & 50.29 & 56.95 & \textcolor{blue}{\textbf{57.71}} \\
Beef & 46.67 & 42.22 & 46.67 & \textcolor{blue}{\textbf{53.33}} & \textcolor{purple}{\underline{48.89}} \\
BeetleFly & \textcolor{purple}{\underline{55.00}} & \textcolor{blue}{\textbf{65.00}} & 50.00 & 58.33 & 53.33 \\
BirdChicken & \textcolor{purple}{\underline{66.67}} & 65.00 & 71.67 & 73.33 & \textcolor{blue}{\textbf{83.33}} \\
BME & \textcolor{blue}{\textbf{55.33}} & \textcolor{purple}{\underline{53.11}} & 43.11 & 52.00 & 49.11 \\
CBF & \textcolor{blue}{\textbf{69.26}} & \textcolor{purple}{\underline{68.89}} & 66.44 & 64.81 & 65.96 \\
Chinatown & 67.83 & 69.00 & \textcolor{blue}{\textbf{70.36}} & 67.44 & \textcolor{purple}{\underline{69.39}} \\
ChlorineConcentration & \textcolor{blue}{\textbf{56.63}} & 48.85 & 53.29 & \textcolor{purple}{\underline{56.30}} & 52.16 \\
Coffee & \textcolor{blue}{\textbf{78.57}} & \textcolor{purple}{\underline{75.00}} & \textcolor{purple}{\underline{75.00}} & \textcolor{purple}{\underline{75.00}} & 73.81 \\
CricketX & 48.72 & \textcolor{purple}{\underline{51.45}} & 50.26 & 49.57 & \textcolor{blue}{\textbf{52.48}} \\
CricketY & 43.59 & \textcolor{purple}{\underline{46.24}} & 46.07 & 44.79 & \textcolor{blue}{\textbf{48.72}} \\
CricketZ & 49.83 & 50.94 & \textcolor{purple}{\underline{51.45}} & \textcolor{blue}{\textbf{52.99}} & 51.88 \\
Crop & 59.01 & 59.35 & \textcolor{blue}{\textbf{59.87}} & \textcolor{purple}{\underline{59.85}} & 59.10 \\
DiatomSizeReduction & \textcolor{blue}{\textbf{86.82}} & 85.51 & \textcolor{purple}{\underline{86.60}} & 85.95 & 85.84 \\
DistalPhalanxOutlineAgeGroup & \textcolor{purple}{\underline{74.82}} & \textcolor{purple}{\underline{74.82}} & \textcolor{blue}{\textbf{75.78}} & 73.14 & 73.14 \\
DistalPhalanxOutlineCorrect & \textcolor{purple}{\underline{75.36}} & \textcolor{purple}{\underline{75.36}} & 72.95 & 71.38 & \textcolor{blue}{\textbf{75.97}} \\
DistalPhalanxTW & \textcolor{purple}{\underline{67.39}} & 65.23 & \textcolor{blue}{\textbf{67.63}} & 66.19 & 63.07 \\
DodgerLoopDay & \textcolor{purple}{\underline{39.58}} & \textcolor{blue}{\textbf{40.42}} & 36.67 & 33.75 & 38.75 \\
DodgerLoopGame & 51.69 & 53.38 & \textcolor{blue}{\textbf{57.00}} & 50.72 & \textcolor{purple}{\underline{53.14}} \\
DodgerLoopWeekend & 85.27 & \textcolor{purple}{\underline{87.20}} & 85.27 & 80.43 & \textcolor{blue}{\textbf{88.65}} \\
Earthquakes & \textcolor{blue}{\textbf{74.10}} & \textcolor{purple}{\underline{73.38}} & 70.74 & 72.66 & 69.78 \\
ECG200 & 84.33 & 83.00 & \textcolor{purple}{\underline{86.33}} & \textcolor{blue}{\textbf{87.67}} & 83.33 \\
ECG5000 & \textcolor{purple}{\underline{93.09}} & 92.39 & 92.71 & \textcolor{blue}{\textbf{93.34}} & 92.39 \\
ECGFiveDays & 66.01 & \textcolor{blue}{\textbf{67.40}} & 66.82 & \textcolor{purple}{\underline{67.48}} & 66.67 \\
ElectricDevices & 51.71 & \textcolor{blue}{\textbf{53.72}} & 48.80 & \textcolor{purple}{\underline{52.54}} & 52.20 \\
FaceAll & 66.69 & 68.32 & \textcolor{blue}{\textbf{68.88}} & \textcolor{purple}{\underline{68.30}} & 62.82 \\
FaceFour & 50.38 & 49.24 & \textcolor{purple}{\underline{52.65}} & \textcolor{blue}{\textbf{58.71}} & 53.79 \\
FacesUCR & 64.59 & 65.61 & \textcolor{blue}{\textbf{67.24}} & \textcolor{purple}{\underline{66.50}} & 65.85 \\
FiftyWords & \textcolor{purple}{\underline{43.37}} & 41.10 & \textcolor{blue}{\textbf{43.74}} & 42.86 & 41.25 \\
Fish & \textcolor{blue}{\textbf{74.67}} & 71.62 & \textcolor{blue}{\textbf{74.67}} & \textcolor{purple}{\underline{74.29}} & 71.62 \\
FordA & 92.58 & \textcolor{blue}{\textbf{92.93}} & 92.17 & \textcolor{purple}{\underline{92.27}} & 91.74 \\
FordB & 76.21 & \textcolor{purple}{\underline{76.63}} & 76.58 & \textcolor{blue}{\textbf{76.79}} & \textcolor{purple}{\underline{76.63}} \\
FreezerRegularTrain & 85.06 & 85.12 & 87.49 & \textcolor{purple}{\underline{90.02}} & \textcolor{blue}{\textbf{91.35}} \\
FreezerSmallTrain & 71.68 & \textcolor{blue}{\textbf{73.64}} & \textcolor{purple}{\underline{73.60}} & 69.65 & 71.37 \\
GestureMidAirD1 & 51.54 & \textcolor{purple}{\underline{52.82}} & 50.51 & \textcolor{blue}{\textbf{54.10}} & 50.51 \\
GestureMidAirD2 & \textcolor{purple}{\underline{37.69}} & 36.67 & \textcolor{blue}{\textbf{41.28}} & 37.18 & 37.44 \\
GestureMidAirD3 & 21.03 & \textcolor{purple}{\underline{22.56}} & \textcolor{purple}{\underline{22.82}} & 22.56 & \textcolor{blue}{\textbf{23.08}} \\
GesturePebbleZ1 & 64.92 & 63.18 & \textcolor{purple}{\underline{67.44}} & 66.86 & \textcolor{blue}{\textbf{67.64}} \\
GesturePebbleZ2 & \textcolor{purple}{\underline{70.04}} & 67.93 & 69.41 & 68.78 & \textcolor{blue}{\textbf{73.63}} \\
GunPoint & 70.44 & 76.89 & \textcolor{blue}{\textbf{81.33}} & \textcolor{blue}{\textbf{81.33}} & \textcolor{purple}{\underline{78.44}} \\
GunPointAgeSpan & \textcolor{purple}{\underline{88.29}} & 85.65 & 86.29 & 88.71 & \textcolor{blue}{\textbf{89.45}} \\
GunPointMaleVersusFemale & 81.22 & \textcolor{purple}{\underline{83.12}} & \textcolor{blue}{\textbf{83.33}} & \textcolor{blue}{\textbf{83.33}} & \textcolor{purple}{\underline{83.23}} \\
GunPointOldVersusYoung & 95.24 & \textcolor{purple}{\underline{97.25}} & \textcolor{blue}{\textbf{99.26}} & 97.14 & 94.39 \\
Ham & 73.33 & 75.24 & \textcolor{blue}{\textbf{76.51}} & \textcolor{purple}{\underline{76.19}} & 74.92 \\
Herring & 58.33 & 58.85 & \textcolor{purple}{\underline{59.38}} & 56.25 & \textcolor{blue}{\textbf{60.42}} \\
InsectWingbeatSound & \textcolor{purple}{\underline{43.59}} & 42.14 & 42.61 & \textcolor{blue}{\textbf{44.49}} & 43.30 \\
ItalyPowerDemand & \textcolor{blue}{\textbf{78.88}} & \textcolor{purple}{\underline{78.78}} & 76.19 & 73.60 & 77.19 \\
Lightning7 & \textcolor{blue}{\textbf{62.10}} & \textcolor{purple}{\underline{60.73}} & 60.27 & \textcolor{blue}{\textbf{62.10}} & \textcolor{blue}{\textbf{62.10}} \\
Meat & 76.67 & 78.33 & \textcolor{purple}{\underline{80.00}} & \textcolor{blue}{\textbf{80.56}} & \textcolor{blue}{\textbf{80.56}} \\
MedicalImages & \textcolor{purple}{\underline{63.33}} & 60.09 & 63.77 & 62.41 & \textcolor{blue}{\textbf{64.39}} \\
MelbournePedestrian & \textcolor{blue}{\textbf{89.72}} & \textcolor{purple}{\underline{88.89}} & 89.11 & 88.33 & 88.88 \\
MiddlePhalanxOutlineAgeGroup & \textcolor{purple}{\underline{58.44}} & 55.41 & 56.28 & \textcolor{blue}{\textbf{59.09}} & 56.28 \\
MiddlePhalanxOutlineCorrect & \textcolor{blue}{\textbf{76.86}} & 73.88 & 76.40 & \textcolor{purple}{\underline{76.75}} & \textcolor{blue}{\textbf{76.86}} \\
MiddlePhalanxTW & 52.60 & 52.60 & \textcolor{blue}{\textbf{56.71}} & \textcolor{purple}{\underline{54.33}} & 53.90 \\
MoteStrain & 63.45 & \textcolor{purple}{\underline{66.77}} & 66.13 & \textcolor{blue}{\textbf{71.35}} & 70.15 \\
OSULeaf & \textcolor{purple}{\underline{65.15}} & 63.09 & \textcolor{blue}{\textbf{65.56}} & 64.60 & 65.29 \\
PhalangesOutlinesCorrect & \textcolor{purple}{\underline{69.31}} & \textcolor{blue}{\textbf{70.98}} & 68.22 & 68.41 & 66.98 \\
PickupGestureWiimoteZ & \textcolor{blue}{\textbf{36.67}} & \textcolor{purple}{\underline{35.33}} & 36.00 & \textcolor{blue}{\textbf{36.67}} & 30.67 \\
Plane & \textcolor{purple}{\underline{91.43}} & 90.16 & \textcolor{blue}{\textbf{92.70}} & \textcolor{purple}{\underline{91.43}} & 90.79 \\
PowerCons & \textcolor{blue}{\textbf{90.56}} & \textcolor{purple}{\underline{90.19}} & 90.37 & 88.70 & 89.63 \\
ProximalPhalanxOutlineAgeGroup & \textcolor{purple}{\underline{85.37}} & \textcolor{purple}{\underline{85.37}} & 83.74 & 85.85 & \textcolor{blue}{\textbf{86.67}} \\
ProximalPhalanxOutlineCorrect & \textcolor{purple}{\underline{79.61}} & 78.47 & 78.92 & \textcolor{blue}{\textbf{79.84}} & 77.78 \\
ProximalPhalanxTW & \textcolor{purple}{\underline{76.10}} & 74.15 & 72.68 & 74.80 & \textcolor{blue}{\textbf{78.05}} \\
ShakeGestureWiimoteZ & 65.33 & 64.00 & 65.33 & \textcolor{purple}{\underline{66.00}} & \textcolor{blue}{\textbf{68.67}} \\
ShapeletSim & \textcolor{blue}{\textbf{77.41}} & 63.52 & \textcolor{purple}{\underline{73.15}} & 71.67 & 70.93 \\
ShapesAll & \textcolor{blue}{\textbf{67.00}} & \textcolor{purple}{\underline{66.11}} & 65.28 & 66.44 & 65.44 \\
SmoothSubspace & \textcolor{blue}{\textbf{95.56}} & \textcolor{purple}{\underline{95.11}} & \textcolor{purple}{\underline{95.11}} & 94.89 & \textcolor{purple}{\underline{95.11}} \\
SonyAIBORobotSurface1 & \textcolor{purple}{\underline{75.10}} & 69.83 & 68.28 & 72.32 & \textcolor{blue}{\textbf{75.93}} \\
SonyAIBORobotSurface2 & \textcolor{purple}{\underline{65.72}} & 65.65 & 65.90 & \textcolor{blue}{\textbf{67.05}} & 65.97 \\
Strawberry & \textcolor{blue}{\textbf{94.14}} & 92.52 & \textcolor{purple}{\underline{93.24}} & 93.33 & \textcolor{purple}{\underline{93.60}} \\
SwedishLeaf & 81.60 & 80.91 & \textcolor{blue}{\textbf{85.01}} & \textcolor{purple}{\underline{82.93}} & \textcolor{purple}{\underline{82.93}} \\
Symbols & \textcolor{purple}{\underline{34.27}} & 32.96 & \textcolor{blue}{\textbf{35.11}} & 34.51 & 35.08 \\
SyntheticControl & 94.89 & \textcolor{blue}{\textbf{95.89}} & \textcolor{purple}{\underline{95.67}} & 94.78 & 94.44 \\
ToeSegmentation1 & 76.61 & \textcolor{purple}{\underline{81.14}} & \textcolor{blue}{\textbf{81.58}} & 74.42 & 56.87 \\
ToeSegmentation2 & 63.33 & \textcolor{blue}{\textbf{69.23}} & 61.54 & \textcolor{purple}{\underline{69.49}} & 66.92 \\
Trace & 90.33 & 86.67 & 88.67 & \textcolor{purple}{\underline{89.33}} & \textcolor{blue}{\textbf{91.67}} \\
TwoLeadECG & 53.38 & 60.90 & 61.02 & \textcolor{purple}{\underline{61.08}} & \textcolor{blue}{\textbf{61.93}} \\
TwoPatterns & 86.75 & \textcolor{blue}{\textbf{88.61}} & \textcolor{purple}{\underline{87.58}} & 85.56 & 87.08 \\
UMD & 68.29 & \textcolor{purple}{\underline{71.99}} & 67.59 & \textcolor{blue}{\textbf{74.31}} & 74.07 \\
UWaveGestureLibraryX & 66.83 & 67.18 & 66.42 & \textcolor{blue}{\textbf{67.97}} & \textcolor{purple}{\underline{67.59}} \\
UWaveGestureLibraryY & 57.44 & \textcolor{blue}{\textbf{58.96}} & \textcolor{purple}{\underline{58.19}} & 58.75 & 58.33 \\
UWaveGestureLibraryZ & \textcolor{purple}{\underline{62.14}} & \textcolor{blue}{\textbf{63.34}} & 60.91 & 62.02 & 60.84 \\
Wafer & \textcolor{purple}{\underline{97.46}} & 97.54 & 97.28 & \textcolor{blue}{\textbf{97.77}} & 97.30 \\
Wine & 51.85 & 58.64 & \textcolor{purple}{\underline{61.11}} & \textcolor{blue}{\textbf{62.96}} & 59.88 \\
WordSynonyms & 35.37 & 36.57 & \textcolor{purple}{\underline{36.94}} & \textcolor{blue}{\textbf{37.10}} & 36.52 \\
Yoga & 78.28 & \textcolor{blue}{\textbf{79.02}} & \textcolor{purple}{\underline{77.67}} & 77.00 & 76.33 \\
\midrule
\textbf{Average} & 67.15 & 67.16 & 67.35 & \textcolor{blue}{\textbf{67.78}} & \textcolor{purple}{\underline{67.48}} \\
\textbf{1$^{st}$ Count} & 19 & 12 & 20 & \textcolor{blue}{\textbf{25}} & \textcolor{purple}{\underline{22}} \\
\bottomrule
\end{tabular}
\label{tab:different_prototypes_ucr_results}
\end{table*}

\vfill

\end{document}